\documentclass[journal]{IEEEtran}
\ifCLASSINFOpdf
\else
\fi

\usepackage{epsfig}
\usepackage{graphicx}
\usepackage{amsmath}
\usepackage{amssymb}

\usepackage{pifont}
\newcommand{\cmark}{\ding{51}}%
\newcommand{\xmark}{\ding{55}}%

\usepackage{makecell}

\begin{document}
%
\title{Cross-Spectrum Dual-Subspace Pairing for RGB-infrared Cross-Modality Person Re-Identification}
%
%
%

\author{Xing Fan, Hao Luo, Chi Zhang, Wei Jiang

\thanks{Xing Fan, Hao Luo, Wei Jiang are with the College of Control Science and Enginneering, Zhejiang University, Hangzhou 310027, China; E-mail: xfanplus@zju.edu.cn, haoluocsc@zju.edu.cn, jiangwei\_zju@zju.edu.cn.}
\thanks{Chi Zhang is with Beijing Research Institute of Megvii Inc (Face++), Beijing, China; E-mail: zhangchi@megvii.com} 

}

%
%

\markboth{Journal of \LaTeX\ Class Files,~Vol.~14, No.~8, August~2015}%
{Shell \MakeLowercase{\textit{et al.}}: Bare Demo of IEEEtran.cls for IEEE Journals}
%



\maketitle

\begin{abstract}
Due to its potential wide applications in video surveillance and other computer vision tasks like tracking, person re-identification (ReID) has become popular and been widely investigated. However, conventional person re-identification can only handle RGB color images, which will fail at dark conditions. Thus RGB-infrared ReID (also known as Infrared-Visible ReID or Visible-Thermal ReID) is proposed.
Apart from appearance discrepancy in traditional ReID caused by illumination, pose variations and viewpoint changes, modality discrepancy produced by cameras of the different spectrum also exists, which makes RGB-infrared ReID more difficult.
To address this problem, we focus on extracting the shared cross-spectrum features of different modalities. In this paper, a novel multi-spectrum image generation method is proposed and the generated samples are utilized to help the network to find discriminative information for re-identifying the same person across modalities. Another challenge of RGB-infrared ReID is that the intra-person (images from the same person) discrepancy is often larger than the inter-person (images from different persons) discrepancy, so a dual-subspace pairing strategy is proposed to alleviate this problem. Combining those two parts together, we also design a one-stream neural network combining the aforementioned methods to extract compact representations of person images, called Cross-spectrum Dual-subspace Pairing (CDP) model. Furthermore, during the training process, we also propose a Dynamic Hard Spectrum Mining method to automatically mine more hard samples from hard spectrum based on current model state to further boost the performance.
Extensive experimental results on two public datasets, SYSU-MM01 with RGB + \textit{near-infrared} images and RegDB with RGB + \textit{far-infrared} images, have demonstrated the efficiency and generality of our proposed method, which outperform the current state-of-the-art approaches by a large margin. Note that, on RegDB dataset, rank-1 accuracy is improved from $50.9\%$ to $65.0\%$, and mAP is improved from $47.0\%$ to $62.7\%$. Our code will be publicly released. 
\end{abstract}


%
\IEEEpeerreviewmaketitle

\section{Introduction}

\begin{figure}[t]
    \begin{center}
        \includegraphics[width=0.8\linewidth]{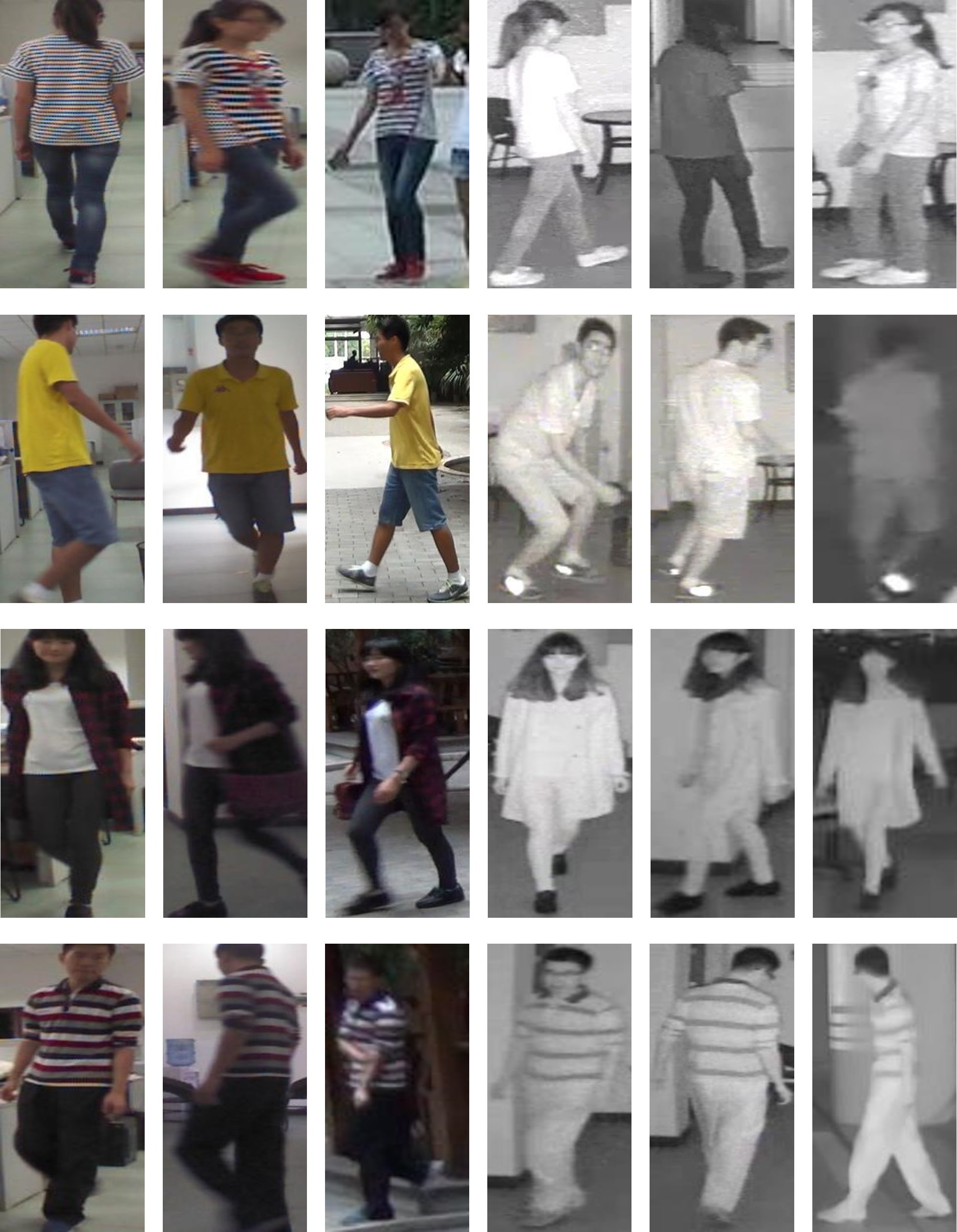}
    \end{center}
    \caption{Sample images from SYSU-MM01~\cite{SYSU-MM01} dataset. Samples of each row belong to the same person. The first three column images are captured by RGB cameras, while the last three column images are captured by infrared cameras.}
    \label{fig:samples}
\end{figure}

Person-identification aims at identifying the same person from different cameras. Given a probe person image, it retrieves images with the same ID in gallery images. It is a difficult problem because the appearance of probe images and gallery images can be quite different even with the identical ID. As shown in Fig.~\ref{fig:samples}, due to illumination, camera model differences, viewpoint changes, and pose variations, the foreground person images can vary a lot. In addition to background clusters, identifying the same person is difficult. Person re-identification is also a zero-shot problem~\cite{DDDM}, which means the IDs in the training set is not used in the testing set, and after learning to identify training IDs, the model needs to discriminate new persons which are never seen before.

Despite those challenges, with the advances of deep neural networks and the continuous effects made by the person re-identification community, RGB ReID has gained a lot of progress and achieved high accuracy~\cite{WhereandWhen,yang2017person,yang2018person,S2S,GLAD_TMM,Triplet_Hard_Loss,PCB+RPP}. 
However, RGB ReID will fail in dark conditions. Without extra artificially lighting, a conventional ReID system using RGB cameras can only work in day time. In other words, the system can not function well in almost half the time of one day, which substantially limits the application of ReID. 

In real-world scenarios, many surveillance cameras will switch to night mode, using an infrared camera to capture images at night. Then a new question arises: can we identify a person across infrared images and RGB images? Compared with RGB images with rich color information and structure patterns, infrared images are captured under the infrared spectrum, resulting in a quite different appearance. As shown in Fig.~\ref{fig:samples}, in the first row the discriminative pattern of the striped T-shirt in RGB spectrum is missing in the infrared spectrum, and in the second row, the distinctive yellow color is lost. The modality discrepancy makes it more difficult to identify the same person between the RGB spectrum and infrared spectrum. This new problem is a cross-modality problem and only a few pioneer works have been made in recent years.

To investigate this RGB-infrared cross-modality person re-identification problem, we need to collect new data including RGB images and infrared images of the same person. \cite{SYSU-MM01} proposes SYSU-MM01 dataset, which contains images captured by 4 RGB cameras and 2 near-infrared cameras of both indoor and outdoor scenarios. A deep zero-padding method is also proposed to extract unified feature vectors for both RGB images and infrared images. The deep zero-padding method is simple and efficient, where RGB sub-network handles RGB images and infrared sub-network handles infrared images. But computation cost for the zero-padded channel is wasted and the shared features of two modalities cannot be well extracted. Instead, in this paper, every input channel is utilized and a cross-spectrum strategy is applied to discover shared features in all modalities.

Apart from one-stream methods, \cite{BDTR} and \cite{HCML} also demonstrate the efficiency of two-stream methods. A Hierarchical Cross-modality Metric Learning (HCML)~\cite{HCML} method first extracts features separately for each modality and then uses shared layers to obtain the final unified features. After that, a metric learning strategy follows to further promote performance. However, a two-stage training process requiring human intervention makes it less convenient for practical large-scale applications. \cite{BDTR} improves this idea and proposes a novel dual-constrained top-ranking loss function to guide the feature learning process, making the training end-to-end. Unlike those pioneer works, we design a one-stream deep convolutional neural network (CNN) to discover shared features with a simpler network structure.

\begin{figure}[t]
	\begin{center}
		\includegraphics[width=0.9\linewidth]{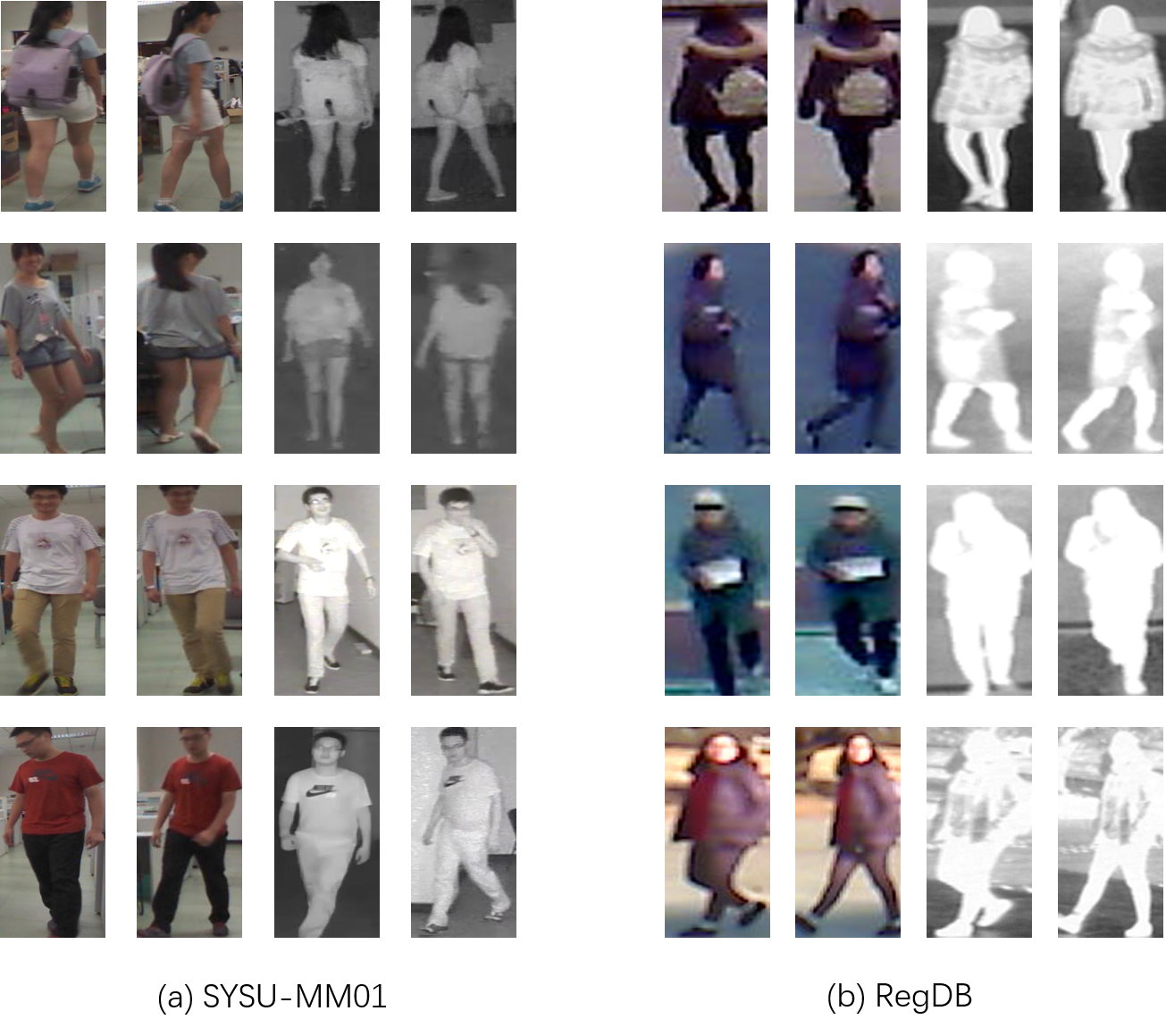}
	\end{center}
	\caption{Comparison of two public datasets. (a) samples from SYSU-MM01~\cite{SYSU-MM01} dataset, and (b) samples from RegDB~\cite{RegDB} dataset. Every four images in a row represent the same person. As we can see, near-infrared images in the SYSU-MM01 dataset and far-infrared images in the RegDB dataset are quite different. A good RGB-infrared method should be robust and can handle both of these scenarios.}
	\label{fig:two-datasets}
\end{figure}

In the HCML~\cite{HCML} work, a new visible-thermal dataset called RegDB is also proposed, which exploits the original data from~\cite{RegDB} and designs an evaluation protocol similar to SYSU-MM01~\cite{SYSU-MM01} dataset. The main difference is, SYSU-MM01~\cite{SYSU-MM01} contains RGB and \textit{\textbf{near-infrared}} images, while RegDB~\cite{RegDB} contains RGB and \textit{\textbf{far-infrared}} images, which makes it even more difficult for a method to work in both two datasets. Some examples are shown in Fig.~\ref{fig:two-datasets}. As we can see, due to its shorter wavelength, near-infrared images from SYSU-MM01~\cite{SYSU-MM01} dataset have a sharp edge and clear background. While, as a comparison, far-infrared images from RegDB~\cite{RegDB} dataset have blurred edges, losing a lot of color patterns. A robust RGB-infrared method needs to handle both of these situations. Our proposed method achieves a new state-of-the-art score in both SYSU-MM01\cite{SYSU-MM01} and RegDB~\cite{RegDB}, demonstrating its efficiency and generality.

Another idea to fill the gap between RGB modality and infrared modality is using a Generative Adversarial Nets (GAN)~\cite{GAN} model. Cross-modality generative adversarial network (cmGAN)~\cite{cmGAN} trains a deep convolutional neural
network as a generator to generate a modality-invariant representation for RGB and IR images in a common subspace and a discriminator to learn a discriminative feature representation from different modalities. Instead of mapping images from different modalities to a common feature subspace, a recent work called Dual-level Discrepancy Reduction Learning ($\text{D}^2\text{RL}$)~\cite{D2RL} translates an infrared image into its
visible counterpart and a visible image to its infrared version using an image-level sub-network to reduce modality discrepancy and uses a feature-level sub-network to reduce the remaining appearance discrepancy through feature embedding. Those works both achieve superior performance, but training a generator and a discriminator makes the network more complex, and how to balance the generator and discriminator is non-trivial. Instead, our new proposed method uses a simpler network structure trained in an end-to-end way.

Although those pioneer works have made great progress in the RGB-infrared ReID task, compared with conventional RGB ReID task, the performance is still inferior, which may limit its wide application in real-world scenarios. The main challenge for RGB-infrared ReID is that the method needs to handle not only appearance discrepancy in traditional ReID caused by illumination, pose variations and viewpoint changes, but also additional modality discrepancy produced by cameras of the different spectrum, which makes RGB-infrared ReID problem more difficult. 

For appearance discrepancy, existing works on traditional RGB ReID task like batch hard triplet loss~\cite{Triplet_Hard_Loss} have been demonstrated that they are efficient to deal with appearance difference and extract discriminative features. We adopt this idea and introduce batch hard triplet loss~\cite{Triplet_Hard_Loss} to supervise our network to push inter-class persons away and pull intra-class persons together. We also add a cross-entropy classification loss to further facilitate the convergence of the training process.

\begin{figure}[t]
    \begin{center}
        \includegraphics[width=0.9\linewidth]{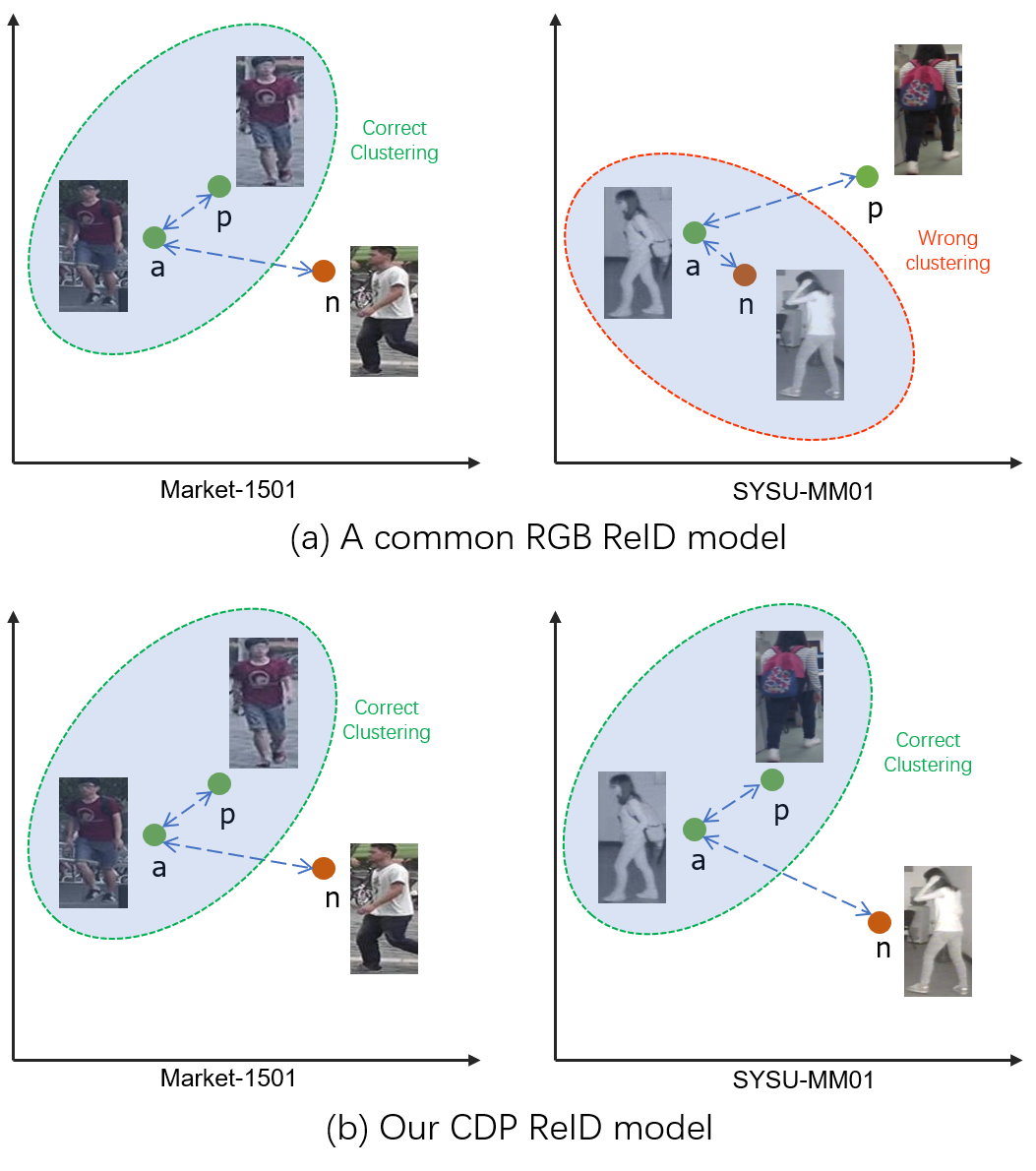}
    \end{center}
    \caption{Comparison of the feature embedding between (a) a common RGB ReID model, and (b) our CDP ReID model. Each dot (green or red) is a feature vector in the embedding space of the input image. $a$, $p$, and $n$ represent anchor, positive, and negative samples respectively. On Market-1501, both models can distinguish different persons successfully, cause images are all from the same modality. On SYSU-MM01, due to modality discrepancy, a common RGB ReID model is more likely to produce a feature feature embedding with $D(a,p)>D(a,n)$, while our CDP ReID model could maintain $D(a,p)<D(a,n)$ well. $D(\cdot)$ is the distance between two embedded features. An actual feature distribution can be seen in Fig.~\ref{fig:vis_feat}.}
    \label{fig:intra-inter}
\end{figure}

As for modality discrepancy, unlike cmGAN~\cite{cmGAN} mapping images of two modalities to common subspace or $\text{D}^2\text{RL}$~\cite{D2RL} mapping images to unified image space, we argue that we can directly extract shared cross-modality features using a one-stream network. To achieve this objective, we need to explicitly guide the network to only focus on the shared features across modalities and ignore the modality-specific features. In this paper, we propose a Cross-Spectrum Image Generation (CSIG) method to generate multi-spectrum images, including the blue spectrum, yellow spectrum, red spectrum, and gray spectrum. As a result, the network is trained with multi-spectrum images and is forced to discover cross-spectrum shared features for correct discrimination of all spectrum.

Another challenge of RGB-infrared ReID is that the intra-person (images from the same person) distance across RGB and infrared cameras is often larger than the
inter-person (images from different persons) distance of the same type of cameras~\cite{D2RL}. As shown in Fig.~\ref{fig:intra-inter}, due to modality discrepancy, images of the same person from two different modalities are quite different, and the distance between anchor image and positive image can be larger than the distance between anchor image and negative image in the embedded space. As shown in the right side of Fig.~\ref{fig:intra-inter}, although anchor and negative images are from different persons, they look quite similar, because they are both infrared images (same modality), while the positive sample is RGB image(cross-modality). To address this problem, we propose a novel dual-subspace pairing strategy which couples the original images from one subspace with a generated image in another subspace. The generated image has an identical underlying appearance with the original image, but in a different modality. In this way, the network can learn to focus on eliminating cross-modality difference and extracting discriminative features from all modalities. Compared with traditional RGB ReID models extracting all-spectrum features, our model focuses on extracting only cross-spectrum features, reducing the modality distraction and producing a more unified feature (an actual feature distribution is shown in Fig.~7).

During the training process of our proposed model, we use the cross spectrum image generation method to produce image pairs. Apart from sampling from all spectrum through a uniform distribution, we could also sampling based on the current model state. Cause with the training progressing, the model can learn to classify samples from one spectrum better than samples from another spectrum. As a result, uniform sampling is not an optimal option. Therefore, we propose a Dynamic Hard Spectrum Mining strategy to automatically evaluate the difficulty level of different spectrum based on the current model state, and assign a harder spectrum a larger sampling probability. The experimental results have demonstrated the efficiency of our proposed method.

The main contribution of this paper can be summarized in four-fold:

(1) Unlike existing RGB-infrared ReID methods, we adopt a one-stream network structure with carefully designed paired input to mining cross-spectrum features, achieving superior performance and providing a new paradigm for future works. More detailed discussion can be found in Section~\ref{sec:discuss-structure};

(2) a cross-spectrum image generation method which can generate multi-spectrum images efficiently with almost no extra computation cost; a dual-subspace pairing strategy that can help the network to learn to reduce the cross-modality discrepancy and extract shared features across modalities;

(3) We also propose a novel Dynamic Hard Spectrum Mining method to automatically mine more hard training samples from hard spectrum based on the current model state during the training process, further promoting the performance of the network;

(4) we design a one-stream neural network with input channel expansion to extract discriminative features for the RGB-infrared cross-modality ReID problem. Combining with the cross-spectrum image generation method and dual-subspace pairing strategy, our network achieves superior performance and outperforms current state-of-the-art methods by a large margin on both SYSU-MM01~\cite{SYSU-MM01} dataset and RegDB~\cite{RegDB} dataset. And our code will be publicly released.

\section{Related Works}

\textbf{Metric Based ReID.} Metric based methods learn a distance metric to determine the similarity between two images. Traditional metric learning algorithms include Locally adaptive Decision Functions (LADF)~\cite{LADF}, Cross-view Quadratic Discriminant Analysis (XQDA)~\cite{LOMO_XQDA}, Data-Driven Distance Metric (DDDM) method~\cite{DDDM}, Keep It Simple and Straightforward Metric Learning (KISSME)~\cite{KISSME} and its improvement Dual-Regularized KISS Metric Learning~\cite{7451215}, Local Fisher Discriminant Analysis (LFDA)~\cite{LFDA} and its kernel variant (k-LFDA)~\cite{k-LFDA}. With the advance of the deep neural network, loss function based metric learning deep methods become popular. \cite{Siamese_CNN} proposes a Siamese Convolutional Neural Network structure (SCNN) and uses the Fisher criterion for deep metric learning. ~\cite{Gated_SCNN} designs a Gated Siamese Convolutional Neural Network (Gated-SCNN) with contrastive loss function proposed in \cite{Contrastive_Loss} as a metric criterion. In the face recognition area, triplet loss function~\cite{Triplet_Loss} has been proven to be efficient for deep metric learning in FaceNet~\cite{FaceNet}. \cite{dingDeepFeatureLearning2015} also introduces the triplet loss function for deep metric learning and achieves good accuracy. After that, ~\cite{Improved_Triplet_Loss} proposes an improved triplet loss function to further promote the performance. But some works~\cite{IDE,chen2017multi} argue that a classification loss, possibly combined with a verification loss, is superior for the ReID task. In defense of triplet loss, ~\cite{Triplet_Hard_Loss} proposes a batch hard triplet loss with a $PK$-sampling batch generation strategy, achieving an impressive performance and becoming popular in ReID task. A variant of triplet loss using an extra sample for distance comparison called Quadruplet Loss~\cite{Quad_Loss} is also proposed, which can lead to the model output with a larger inter-class variation and a smaller intra-class variation compared to the triplet loss. Apart from point-to-point metric, Set-to-set (S2S) distance metric~\cite{S2S} is also investigated, in which the underline objective focuses on preserving the compactness of intraclass samples for each camera view, while maximizing the margin between the intraclass set and interclass set. Multilinear Multi-index Fusion method~\cite{Multi-index} explores the complementary information hidden in different indexes more thoroughly by formulating this procedure as a multilinear based optimization problem. 
Re-ranking method is also widely used in the ReID task, such as k-Reciprocal re-ranking~\cite{Re-ranking}, Ranking Aggregation of Similarity Pulling and Dissimilarity Pushing~\cite{Ranking-agg}, and Reciprocal Optimization (RO) method~\cite{RO-re-ranking}.

\textbf{Part Based ReID.} Pose is an important clue to locate different human body parts, thus many pose-based ReID methods are proposed, which divide the input image into different parts using pose information. 
Pose-driven Deep Convolutional (PDC)~\cite{PDC} model explicitly leverages the human part cues and transfers both global human body and local body parts into
normalized and homologous state for better feature embedding.
Spindle Net~\cite{SpindleNet} adopts a human body region guided multi-stage feature decomposition and tree-structured competitive feature fusion. The Global-Local-Alignment Descriptor (GLAD)~\cite{GLAD_TMM} estimates four human key points that are robust for various poses and camera viewpoints using an external model and then extracts corresponding features of the global body and local parts.
PoseBox~\cite{PoseBox}  generates a PoseBox structure through pose estimation followed by affine transformations to align pedestrians to a standard pose and reduce the impact of pose estimation errors and information loss during PoseBox construction through a PoseBox fusion CNN architecture.
Pose-Normalized GAN (PN-GAN)~\cite{PN-GAN} designs a specific generative adversarial network for pose normalization, and the model can learn a new type of deep ReID features free of the influence of pose variations using synthesized images, which are complementary to features learned with the original images.
Feature Distilling GAN (FD-GAN)~\cite{FD-GAN} also uses GAN to learn identity-related and pose-unrelated representations based on a Siamese structure with multiple novel discriminators on human poses and identities. Apart from pose-based part methods, ~\cite{Improved_Triplet_Loss,Siamese-LSTM,PCB+RPP} adopt a part-division strategy using horizontal stripes, which is simple and demonstrated to be efficient. ~\cite{latent_parts} adds three novel constraints on  Spatial Transformer Networks (STN)~\cite{STN} for effective latent parts localization and body-part feature representation, then fuses full-body and body-part identity discriminative features for powerful pedestrian representation. ~\cite{cluster_parts} gathers together feature maps depicting different body parts by clustering feature maps based on the location of their maximum responses. 

\textbf{Other RGB ReID.} ~\cite{matsukawaPersonReidentificationUsing2016} leverages additional attribute data and proposes new labels by combining different attribute labels for an additional classification loss function. ~\cite{Siamese-LSTM} presents a novel siamese Long Short-Term Memory (LSTM) architecture that can process image regions sequentially and enhance the discriminative capability of local feature representation by leveraging contextual information.
~\cite{xiaoLearningDeepFeature2016} proposes a Domain Guided Dropout method that assigns each neuron a specific dropout rate for each domain according to its effectiveness on that domain. ~\cite{Mask-Guided} introduces the binary segmentation masks to construct synthetic RGB-Mask pairs as network inputs and design a mask-guided contrastive attention model (MGCAM) to learn features from the body and
background regions separately. And~\cite{Mutual} presents a deep
mutual learning (DML) strategy based on model distillation where, rather than a one-way transfer between a static predefined teacher and a student, an ensemble of students learn collaboratively and teach each other throughout the training process.
~\cite{SGGNN} proposes  a novel deep learning framework, called
Similarity-Guided Graph Neural Network (SGGNN) to overcome the limitation that the similarity estimation of some hard samples might not be accurate.
Apart from image-based ReID, video-based ReID is also an important task where both temporal and spatial information in the videos need to be handled properly. For example, Deep Siamese Attention Networks~\cite{WhereandWhen} designs a siamese attention architecture that jointly learns spatiotemporal video representations and their similarity metrics. Data labeling is also time-consuming for video-based ReID, and Feature Affinity-based Pseudo Labeling (FAPL) method~\cite{FAPL} employs pseudo-labeling by measuring the affinity of unlabeled samples with the underlying clusters of labeled data samples using the intermediate features of deep neural networks. ReID can also be integrated into the person tracking task. Context-Aware Hypergraph Modeling~\cite{Context-Aware-Hypergraph} uses a hypergraph representation to link related objects for re-identification and a diverse hypergraph ranking technique for person-focused network summarization.

\textbf{RGB-infrared ReID.} Compared with conventional RGB ReID, the RGB-infrared ReID is a relatively new problem and it needs to address not only the appearance discrepancy but also the modality discrepancy. Although it is a challenging task, some great pioneer works have been made. ~\cite{SYSU-MM01} contributes a new multiple modality Re-ID dataset named SYSU-MM01, including RGB and IR images, and proposes a deep zero-padding method to train a one-stream network towards automatically evolving domain-specific nodes in the network for cross-modality matching. ~\cite{cmGAN} designs a novel cross-modality generative adversarial network (cmGAN) to learn discriminative feature representation from different modalities. ~\cite{HCML} introduces a two-stage framework including a TwO-stream CNN NEtwork (TONE) to learn the multi-modality sharable feature representations and a Hierarchical Cross-modality Metric Learning (HCML) method. However, a two-stage process requires a human intervention, which limits its wide large-scale application in real-world scenarios. In another work, ~\cite{BDTR} proposes a dual-path network with a novel bi-directional dual-constrained top-ranking loss
to learn discriminative feature representations in an end-to-end way. Recently, ~\cite{D2RL} introduces a novel Dual-level Discrepancy Reduction Learning ($\text{D}^2\text{RL}$) scheme which reduces modality discrepancy using an image-level sub-network and reduces the remaining appearance discrepancy through a feature-level sub-network. Hypersphere Manifold Embedding (HSME)~\cite{D-HSME} maps the input images onto a hypersphere manifold, and acquires decorrelated features with a two-stage training scheme.

\section{Our Approach}

In this section, we will describe the proposed cross-spectrum image generation method and dual-sub-space pairing strategy in details, then we will introduce our designed one-stream network for feature extracting and its corresponding loss function we adopted. Finally, the dynamic hard spectrum mining strategy is introduced, and a discussion of different network structures follows.

\subsection{Cross-spectrum Image Generation}

\begin{figure}[t]
    \begin{center}
        \includegraphics[width=0.85\linewidth]{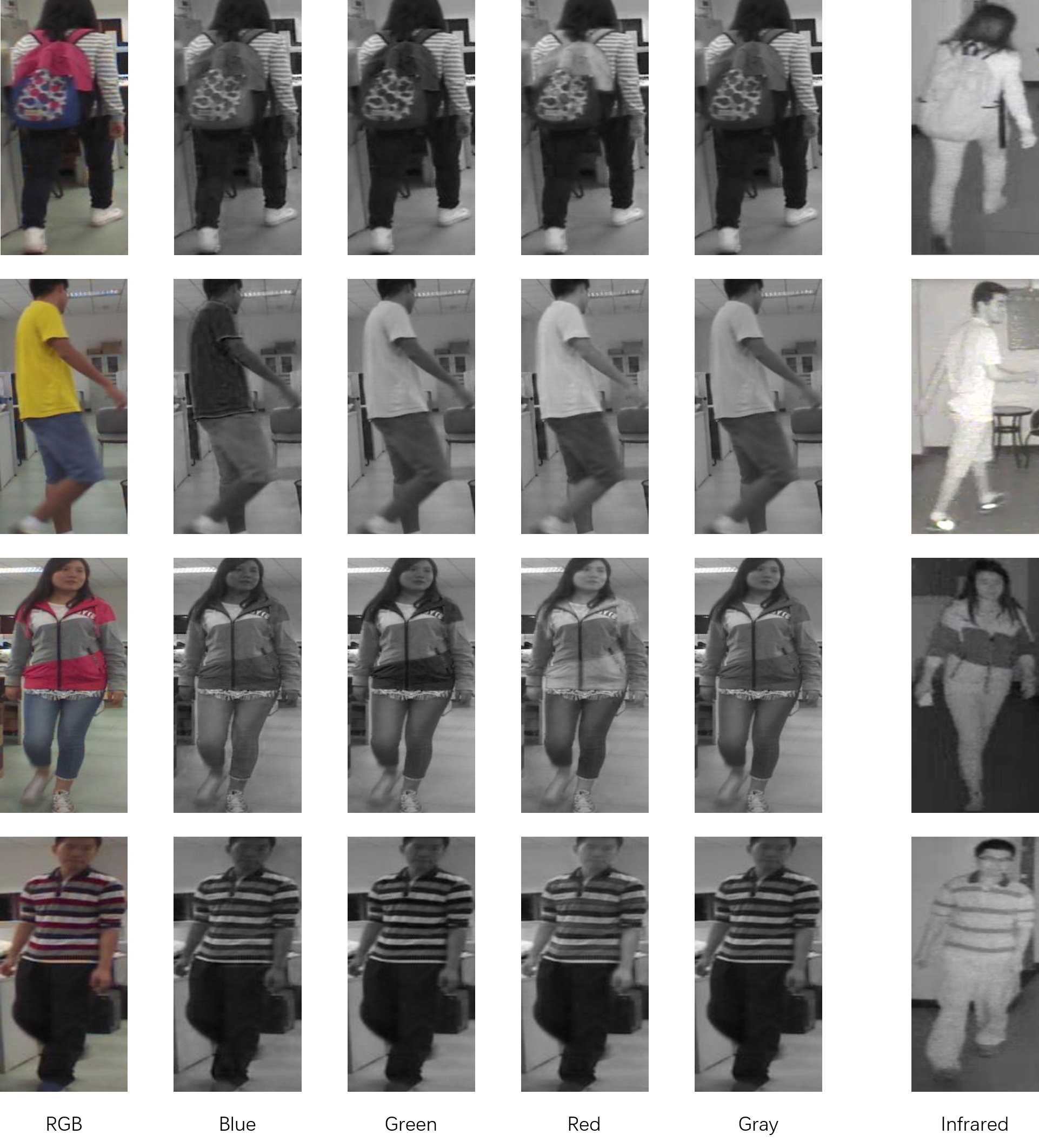}
    \end{center}
    \caption{Examples of our cross-spectrum generation image method. The first column is the original RGB images, the second to the fifth columns are generated images in the blue spectrum, green spectrum, red spectrum, and gray spectrum respectively. And the last columns are similar reference images of the same person captured by an infrared camera.}
    \label{fig:generation}
\end{figure}

The core challenge for RGB-infrared cross-modality ReID is that the same person with exactly the same appearance will produce a quite different image under RGB cameras and infrared cameras, which we call modality-discrepancy.

Modality-discrepancy is caused by the spectrum difference that RGB cameras capture the features of a person in the RGB spectrum while infrared cameras capture the features in the infrared spectrum. From one modality to another modality, some information is irreversibly lost and can not be recovered. As we can see in Fig.~\ref{fig:samples}, the yellow color of the cloth and striped pattern of the T-shirt in RGB images are missing in infrared images. Based on this observation, we focus our attention on extracting features existing in all modalities, in other words, the shared cross-modality features.

To accomplish this goal, we can directly feed RGB images and infrared images together into a one-stream network and let the network learn to extract unified features automatically, no matter which modality the input image is. 
However, without explicit constraints, the network can only achieve an inferior performance, which is verified from the experimental results in Section~\ref{Experiments}.

Instead, we propose a novel cross-spectrum image generation method which can produce images in multi-spectrum. Given an RGB image, we produce images in the red spectrum, green spectrum, and blue spectrum by extracting pixels from the corresponding color channel. In addition, we also add the gray-scale image of the original RGB image as a special spectrum to further increase the number of modalities. For an input RGB color image $C$ with red channel $R$, green channel $G$, blue channel $B$ and corresponding gray-scale image $X$, our cross-spectrum image generation method can be denoted as follows:

\begin{equation}
C~{\stackrel {f}{\to }}~\{R,G,B,X\}
\end{equation}

\noindent where $f$ is our cross-spectrum image generation function.

Some examples are shown in Fig.~\ref{fig:generation}. We can observe that the four generated images in different spectrum show kind of diverse appearances, and are similar to the appearance captured by an infrared camera. Note the bi-color pattern backpack in the first row, this pattern disappears in the green spectrum image, as well as in the infrared spectrum image.

Introducing more modalities into the training process can add extra supervision signal for the network to discover discriminative shared features existing in all modalities, in other words, cross-spectrum shared features, leading to a better feature embedding. 

To maintain consistency, for infrared images, we simply apply a random brightness jitter augmentation with a fixed jitter ratio $\delta$.

Experimental results in Section~\ref{Experiments} demonstrate the efficiency of the proposed cross-spectrum image generation method.

\subsection{Dual-subspace Pairing Strategy}

\begin{figure}[ht]
    \begin{center}
        \includegraphics[width=0.9\linewidth]{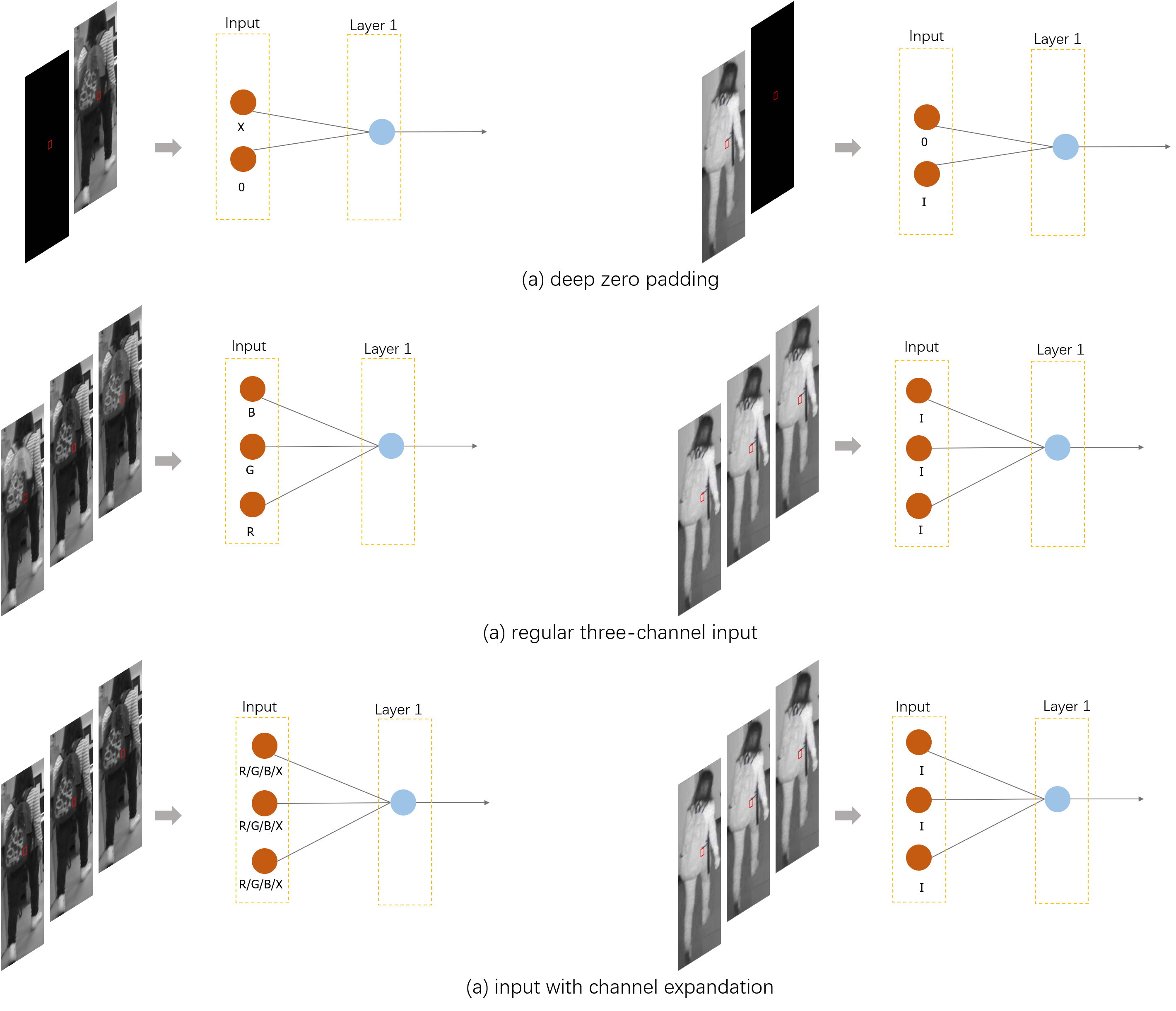}
    \end{center}
    \caption{Comparison of three different input types for one-stream networks. Values from an input position indicated by a red box and a node from layer 1 are chosen for illustration purpose. $R$, $G$, $B$, $X$, and $I$ represent values from red, green, blue, gray and infrared image respectively. Zero-padded channels are all 0 values. The left side is input from RGB modality, while the right side is input from infrared modality.}
    \label{fig:comparison}
\end{figure}

\begin{figure*}[ht]
    \begin{center}
        \includegraphics[width=0.9\linewidth]{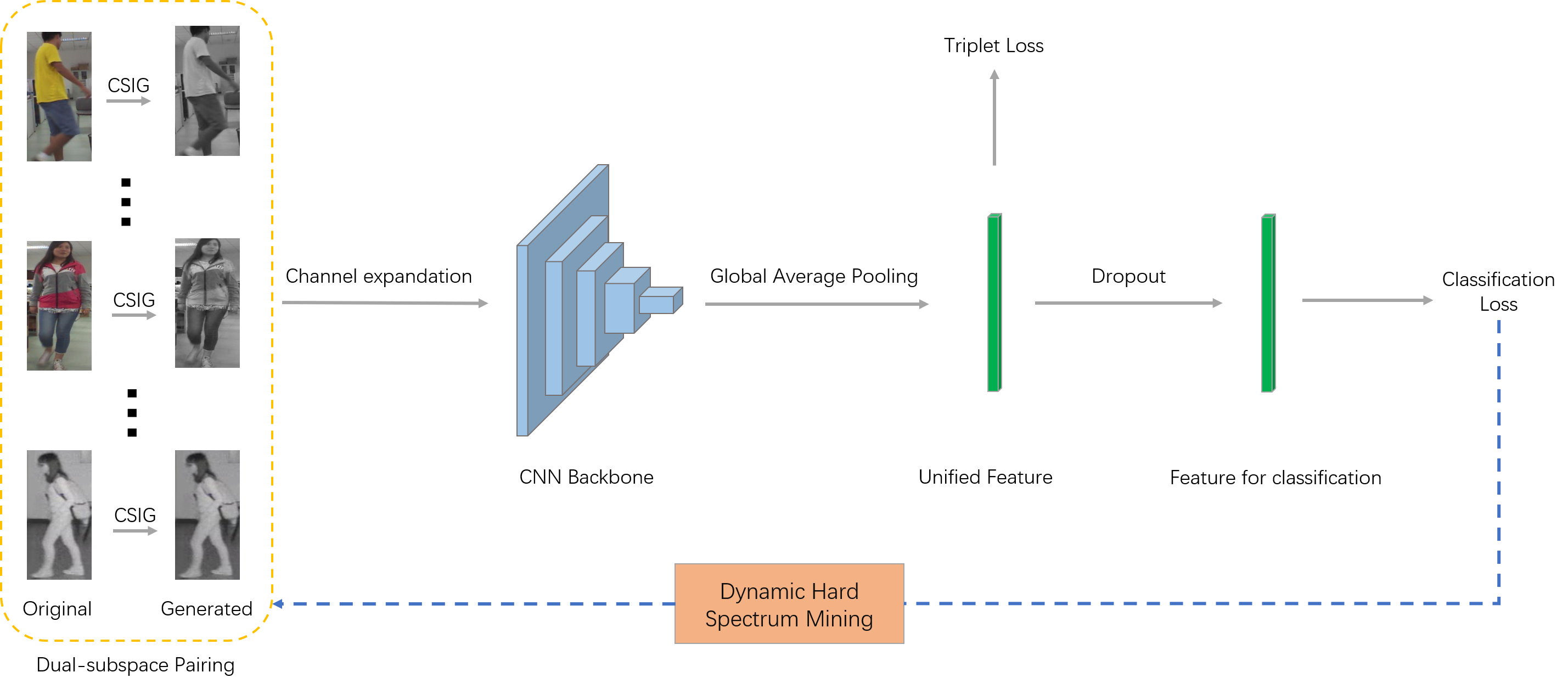}
    \end{center}
    \caption{The overall structure of our proposed approach. We first use the cross-spectrum image generation method to generate images in a different spectrum. A dual-subspace pairing strategy is adopted to pair each original image with a generated image. After a channel expansion, images are fed into a one-stream CNN backbone for feature extracting. The final unified features for both modalities are obtained through a global average pooling applied on the CNN backbone output. Classification loss and triplet loss are used to supervise the training process. Dropout is also introduced to prevent over-fitting and improve robustness. During the training process, a dynamic hard spectrum mining strategy is proposed to further promote the performance.}
    \label{fig:network}
\end{figure*}

Another challenge of RGB-infrared cross-modality ReID is that the intra-person distance across RGB and infrared cameras is often larger than the
inter-person distance of the same type of cameras. As shown in Fig.~\ref{fig:intra-inter}, in traditional RGB ReID dataset like Market-1501~\cite{Market-1501}, images of the same person are more visually similar to the images of different persons. While in an RGB-infrared dataset like SYSU-MM01~\cite{SYSU-MM01}, images of the same person in two different modalities are quite dissimilar due to modality discrepancy, and the dissimilarity can be even larger than which lies between different persons within the same modality.

As a result, in the training process, the network needs to learn to classify samples with smaller distance into different classes and classify samples with larger distance into the same class, which is confusing and hard to converge, leading to inferior performance.

To address this problem, we propose a novel dual-subspace pairing strategy. As shown in Fig.~\ref{fig:network}. For each input original image, we use a cross-spectrum image generation method to produce a paired image, and then feed them together into the network.

Through cross-spectrum image generation, the underlying actual appearance is exactly the same between the original image and the generated paired image. Therefore, the appearance discrepancy can be neglected and ideally the embedded cross-spectrum shared feature should be the same for both the original and generated images.

In this dual-subspace pairing way, the network can focus on learning to eliminate the cross-modality discrepancy between two subspaces and extract discriminative cross-spectrum features.

Specifically, for each input original image, we randomly choose one cross-spectrum image from all generated images $\{R,G,B,X\}$ for RGB image and the random brightness jittered image for infrared image, as the paired counterpart.

\subsection{Our One-Stream Network}

We also design a one-stream deep convolutional neural network for an efficient feature embedding.

First, for all input original images, we apply our proposed cross-spectrum image generation method to create coupled dual-subspace pairs. However, the generated images have only one channel as infrared images, while RGB images have three color channels. We introduce a channel expansion strategy to expand the single channel image to three channels by duplicating, so all input images will have three channels.

As shown in Fig.~\ref{fig:comparison}, three possible input types can be used for a one-stream network:

(a) The deep zero-padding way first converts RGB color image into a gray-scale image, then adds a zero-channel for both gray image and infrared images, producing a two-channel input. For zero-padded gray channel or infrared channel, some nodes will output zero constantly (not activated), which means they are domain-specific. Therefore, gray-scale sub-network and infrared sub-network can handle each modality separately, and the shared nodes can extract shared features for both modalities. However, the computation cost for zero-channel is wasted, and there no explicit constraint for digging shared features across modalities.

(b) We can also simply use a three-channel input for RGB images, and expand channels of infrared images. In this way, nodes need to learn to extract features from both the RGB image and infrared images.

(c) With our cross-spectrum image generation method, we can also choose one generated image in the red, green, blue, or gray spectrum, and expand the one-channel image into a three channel input. Compared with the regular three-channel input that three weights of a node in layer 1 handle only one color channel and one infrared channel respectively, for the input with channel expansion, every weight of a node in layer 1 needs to handle all red, green, blue, gray, and infrared channels, which forces the network to discover discriminative cross-spectrum features existing in all spectrum. Experiments in Section~\ref{Experiments} demonstrate that those cross-spectrum features can indeed help promote the performance of cross-modality ReID.

Following ~\cite{Triplet_Hard_Loss}, we use a $PK$ sampling strategy to construct an input mini-batch, randomly choosing $P$ persons and $K$ images for each person. With the dual-subspace pairing strategy, we generate a paired image using cross-spectrum image generation method for each original image, resulting in a final mini-batch size of $2PK$. 

We choose a standard ResNet-50~\cite{ResNet} structure as our CNN backbone to extract feature maps of input images after the last convolutional layer. Then, as shown in Fig.~\ref{fig:network}, a global average pooling layer is applied to obtain an embedded unified feature vector. We then introduce a dropout layer to prevent over-fitting and produce more discriminative features, followed by a classification fully-connected layer and a softmax function to predict the probability distribution $p_{i,j}$ of $i$-th input image and $j$-th predicted class.  A classification cross-entropy loss is computed based on the predicted probability distribution, which can be formulated as follows:

\begin{equation}
\begin{split}
L_{cls}=-\frac{1}{2PKM}\sum_{i=1}^{2PK}\sum_{j=1}^{M}y_{i,j}\log(p_{i,j})
\end{split} 
\end{equation}

where $M$ is the total number of training persons, $y_{i,j}=1$ when $j=t$ is the target class and $y_{i,j}=0$ otherwise.

We use a batch hard triplet loss on the original image set to further maximize inter-person distance and reduce the intra-person discrepancy. The generated image set of multi-spectrum is quite different, and a batch hard triplet loss on the generated set turns to be hard to converge, so we only apply the batch hard triplet loss on the original image set. The batch hard triplet loss can be formulated as follows:

\begin{equation}
\begin{split}
L_{tri}=\frac{1}{PK}\sum_{p=1}^{P}\sum_{k=1}^{K}\biggl[ m&+\max_{i=1...K}{D(x_p^k,x_p^i)}\ \\
&-\min_{\substack{n=1...P\\j=1...K\\n\ne p}}{D(x_p^k,x_n^j)}\biggr]_+
\end{split} 
\end{equation} 

\noindent where $m$ is a margin constant, $[x]_+$ is $\max(0,x)$ function to guarantee the non-negativity constraint, and $D(\cdot)$ is a distance function between two
features, which we use Euclidean distance in this work.

The total loss of our network is the sum of classification and batch hard triplet loss:

\begin{equation}
L=L_{cls}+\lambda L_{tri}
\end{equation}

\noindent where $\lambda$ is the trade-off between classification loss and triplet loss. To keep simplicity, we fix $\lambda$ to 1 in all our experiments.

\subsection{Dynamic Hard Spectrum Mining}
In the cross-spectrum image generation process, every candidate spectrum is treated equally, and gray, blue, red, yellow colors have the same probability to be chosen, which is not optimal. During the training process, the model learns to extract cross-spectrum discriminate features from each spectrum. Some spectrum can be learned fast, and some other spectrum can be learned slow. It is hard to manually determine which spectrum is easy or hard to learn, and the difficulty level may vary due to different model structures or parameter settings. Therefore, we also propose a Dynamic Hard Spectrum Mining (DHSM) strategy to automatically evaluate the difficulty level of different spectrum, and assign harder spectrum a larger sampling probability.

In this way, once the model learns to handle a specific spectrum, the sampling probability decreases, and there is less training data from this spectrum in the future. As discussed in ~\cite{Triplet_Hard_Loss}, being told over and over again that an easy sample belongs to some specific person I already know does not teach one anything, we need to be presented with hard samples which we currently can not handle to learn new knowledge. So easy samples should be reduced and hard samples should be mined. Unlike traditional sample-level hard mining like Triplet Hard~\cite{Triplet_Hard_Loss} or Quadruplet Hard~\cite{Quad_Loss}, in this paper, we propose a novel spectrum-level hard mining method to dynamically mine hard spectrum during the training process.

Given the $i$-th input cross-spectrum generated image in a training epoch, the current predicted classification probability of target class is $p^i$, and $S(p^i)$ is the spectrum of this sample, which could be R, G, B, and X in this paper. The overall confidence of a specific spectrum during a training epoch can be defined as follows:

\begin{equation}
R_q=\frac{1}{|{p^i|i=1...N_t,S(p^i)=q}|}\sum_{\substack{i=1\\S(p^i)=q}}^{N_t}{p^i}, q=\{R,G,B,X\}
\end{equation}

\noindent where $N_t$ is the total number of generated images in the last training epoch, $|\cdot|$ is the number of samples. Now, the sampling probability of each spectrum in the $t$-th epoch is computed as:

\begin{equation}
\label{equ:Pt}
P_q^t=\frac{1-R_q}{\sum_{q\in \{R,G,B,X\}}{1-R_q}}, q=\{R,G,B,X\}
\end{equation}

As shown in Equation~\ref{equ:Pt}, if the model predicts images from a specific spectrum with a high target probability $R_q$, then the new sampling probability $P_q^t$ is small, meaning there are less generated images of this spectrum in the next training epoch. Through a normalization operation, the total probability is $\sum_{q\in\{R,G,B,X\}}P_q^t=1$. To further utilize the historical information and keep the sampling probability smoothly changing, the final sampling probability is obtained by:

\begin{equation}
\hat{P}_q^{t+1}=\hat{P}_q^t\times\alpha+P_q^t\times(1-\alpha), q=\{R,G,B,X\}
\end{equation}

\noindent where $\hat{P_q^t}$ is the actual sampling probability for $t$-th epoch, and $\alpha$ is a constant value to balance the historical probability and current probability. Now, the cross-spectrum image-generation with dynamic hard spectrum mining can be formulated as follows:

\begin{equation}
C~{\stackrel {f_{\hat{\mathbb {P}}^t}}{\to }}~\{R,G,B,X\}
\end{equation}

\noindent where $\hat{\mathbb {P}}^t(X=q)=\hat{P}_q^t,q\in\{R,G,B,X\}$ is the current sampling probability distribution. With this dynamic hard spectrum mining strategy, we could mine hard samples from the hard spectrum as shown in Fig.~\ref{fig:network}, benefiting to the training process and further promoting the model performance. 

\subsection{Discussion of Network Structures}
\label{sec:discuss-structure}

\begin{figure}[ht]
	\begin{center}
		\includegraphics[width=0.9\linewidth]{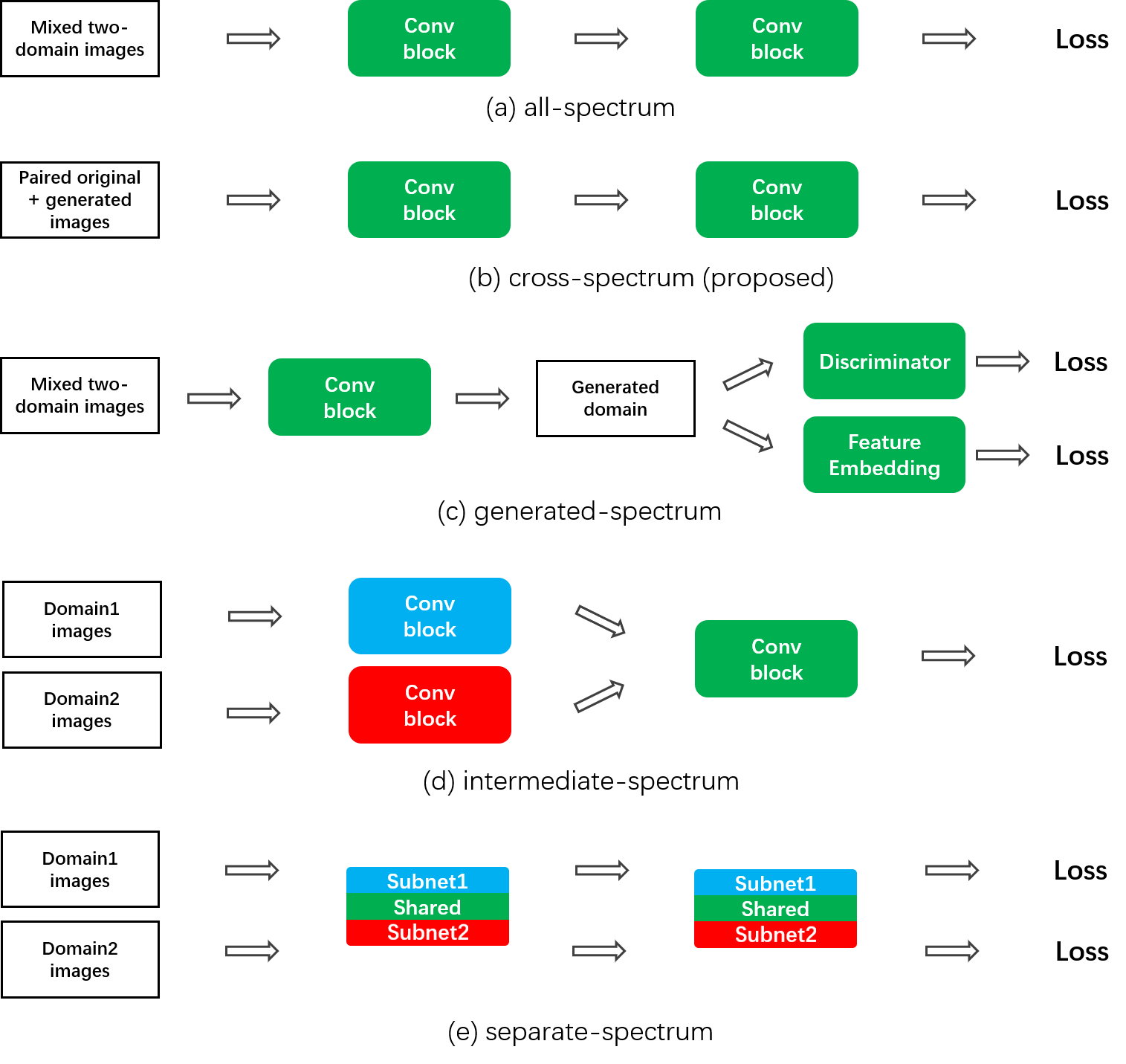}
	\end{center}
	\caption{Comparison of different network structures. Green blocks are shared for both domains, while blue and red blocks are domain-specific. Our proposed network structure is different from existing RGB-infrared ReID methods. More discussion can be found in Section~\ref{sec:discuss-structure}}
	\label{fig:different-spectrum}
\end{figure}

In the pioneering zero-padding~\cite{SYSU-MM01} RGB-infrared cross-modality ReID work, four different network structures are categorized and compared. After that, some other RGB-infrared cross-modality ReID methods have been proposed, and now we could revise which structure the methods for this task actually use and what is their difference. Unlike~\cite{SYSU-MM01}, in this paper, we categorize common RGB-infrared ReID network structures into 5 types. All structures are shown in Fig.~\ref{fig:different-spectrum}, and we will discuss these structures respectively.

\textbf{All-Spectrum} Traditional RGB ReID methods like SVDNet~\cite{sunSVDNetPedestrianRetrieval2017} and Triplet Hard~\cite{Triplet_Hard_Loss} usually use a one stream network structure. As shown in Fig.~\ref{fig:different-spectrum} (a), we can directly apply this structure into the RGB-infrared ReID task with a mixed two-domain input. Cause there is only one domain in the original RGB ReID task, in this way, RGB images and infrared images are treated the same, and features from all spectrum are extracted, resulting in an inferior performance. This structure can be used as a baseline network, just like in zero-padding~\cite{SYSU-MM01}, $\text{D}^2\text{RL}$~\cite{D2RL}, and this paper.

\textbf{Cross-Spectrum} As shown in Fig.~\ref{fig:different-spectrum} (b), the key difference between our proposed cross-spectrum method and all-spectrum methods is that, we train the network with a carefully designed input through the cross-spectrum dual-subspace strategy, which can force the network to discover cross-spectrum features and ignore spectrum-specific features. Experimental results on both SYSU-MM0~\cite{SYSU-MM01} dataset and RegDB~\cite{RegDB} dataset have demonstrated the efficiency and generality of the proposed method, outperforming the state-of-the-art methods by a large margin. Our structure is quite different from existing RGB-infrared methods, and provide a new paradigm for the future works. Furthermore, our structure is similar to the traditional RGB ReID network, hence we can benefit from progress in the RGB ReID task and incorporate it into our network to promote the performance.

\textbf{Generated-Spectrum} Since RGB images and infrared images are from different spectrum, if we want to conduct cross-modality searching, it is reasonable to map those two domain inputs into a generated common domain first. In cmGAN~\cite{cmGAN}, a generator is used to map an image into a feature vector within a shared common space (generated-spectrum), then feature embedding and discriminator follow, as shown in Fig.~\ref{fig:different-spectrum} (c). $\text{D}^2\text{RL}$~\cite{D2RL} also applies a similar network structure, but unlike cmGAN~\cite{cmGAN} using a feature vector, it generates a four-channel image in a unified space to reduce the image-level discrepancy.

\textbf{Intermediate-Spectrum} Low-level features are different for different domain images, but the high-level features for the same person should be consistent, therefore a two-stage network structure could be used to first capture domain-specific features separately, obtaining intermediate-spectrum representations, and then extract domain-shared features through a shared module in the second stage, as shown in Fig.~\ref{fig:different-spectrum} (d). HCML~\cite{HCML} adopts this network structure and optimize the model with a cross-modality metric learning method. BDTR~\cite{BDTR} employs this structure with a dual-constrained top-ranking loss. HSME~\cite{D-HSME} also utilize this structure to learn a hypersphere manifold embedding.

\textbf{Separate-Spectrum} Zero-padding~\cite{SYSU-MM01} uses a zero-value padded four-channel input. Since the padded input channels are all zero, the activation of the padded input is also zero (omitting bias). In this zero-padding way, the one-stream network can act like a two-stream network, automatically evolving domain-specific nodes in the network for cross-modality matching, as shown in Fig.~\ref{fig:different-spectrum} (e). But separating two domains into two different sub-network paths without cross-spectrum feature mining hinders the improvement of performance.

\section{Experiments}
\label{Experiments}

\subsection{Datasets}
\textbf{SYSU-MM01}~\cite{SYSU-MM01} is a large-scale benchmark dataset for RGB-infrared cross-modality ReID. The images are captured by six cameras, including four RGB cameras and two infrared cameras. Three cameras are indoor scenes and the other three cameras are outdoor scenes, which makes it more challenging. Overall, there are 491 identities with 287,628 RGB images and 15,792 infrared images. The dataset has a fixed partition of the training set and testing. For training, a total of 32,451 images including 19,659 RGB images and 12,792 infrared images are used. For testing, we follow the standard evaluation protocol proposed by the dataset author~\cite{SYSU-MM01}. Since there are multiple groundtruths in the gallery images, standard cumulated matching characteristics (CMC) and
mean average precision (mAP) are adopted as an evaluation metric to measure the performance. Single-shot all-search scores computed by the SYSU-MM01~\cite{SYSU-MM01} evaluation code are reported.

\textbf{RegDB}~\cite{RegDB} is a visible-thermal person re-identification dataset captured by visible and far-infrared cameras. It contains 412 persons, and each person has 10 visible images and 10 thermal images. We follow the evaluation protocol in~\cite{HCML}, which randomly split the dataset into two halves, one for training and the other for testing. For testing, images of one modality act as query and images of the other modality act as the gallery. Both thermal-to-visible and visible-to-thermal results of CMC and mAP are computed. As in~\cite{HCML}, this procedure is repeated 10 times to achieve statistically stable results, and the average scores are reported.

\subsection{Implementation Details}
We implement our proposed method using PyTorch deep learning framework. The input image size is $256\times 128$, random horizontal flipping data augmentation is applied. We initialize our ResNet-50~\cite{ResNet} CNN backbone with weights pre-trained on ImageNet. Adam optimizer~\cite{Adam} with the default hyper-parameter values ($\epsilon=10^{-3}$, $\beta_1=0.9$, $\beta_2=0.999$) is used. The random brightness jitter ratio $\delta$ for the infrared image is 0.1. The initial learning rate is $10^{-3}$ and then reduced to $10^{-4}$ and $10^{-5}$. The dropout rate is set to 0.5. $\alpha$ is set to 0.1.
Margin value $m$ for batch hard triplet loss is fixed to 0.3. For SYSU-MM01~\cite{SYSU-MM01}, $PK$ sampling parameters are $P=16$ and $K=4$, thus the original image number in a mini-batch is 64. With a paired image for each original image, the final batch size is 128. The overall training phase is set to 200 epochs. For RegDB~\cite{RegDB}, since it contains fewer samples, we train the network only for 100 epochs and keep the rest of settings identical to SYSU-MM01~\cite{SYSU-MM01} model. More implementation details can be found in the released code.

\subsection{Comparison with the State-of-the-art Methods}
To evaluate the efficiency of our proposed method, we compare the performance of our best Cross-spectrum Dual-subspace Pairing (CDP) model with state-of-the-art methods on two public datasets.

\begin{table}\footnotesize
    \caption{Comparison of our proposed method and the state-of-the-art methods on SYSU-MM01~\cite{SYSU-MM01} dataset.}
    \label{tab:SOTA-SYSU}
    \begin{center}
        \begin{tabular}{c|c|c c c|c}
            \Xhline{2\arrayrulewidth}
            Method & Reference & r=1 & r=10 & r=20 & mAP\\
            \hline
            One-stream~\cite{SYSU-MM01}  &ICCV 2017 & 12.0 & 49.7 & 66.7 & 13.7 \\
            Two-stream~\cite{SYSU-MM01}  &ICCV 2017 & 11.7 & 48.0 & 65.5 & 12.9 \\
            Zero-Padding~\cite{SYSU-MM01}&ICCV 2017 & 14.8 & 54.1 & 71.3 & 16.0 \\
            TONE~\cite{HCML,BDTR}        &AAAI 2018 & 12.5 & 50.7 & 68.6 & 14.4 \\
            HCML~\cite{HCML,BDTR}        &AAAI 2018 & 14.3 & 53.2 & 69.2 & 16.2 \\
            BDTR~\cite{BDTR}             &IJCAI 2018& 17.0 & 55.4 & 72.0 & 19.7 \\
            cmGAN~\cite{cmGAN}           &IJCAI 2018& 27.0 & 67.5 & 80.6 & 27.8 \\
            $\text{D}^2\text{RL}$~\cite{D2RL} &CVPR 2019 & 28.9 & 70.6 & 82.4 & 29.2 \\
            D-HSME~\cite{D-HSME}         &AAAI 2019 & 20.7 & 62.7 & 78.0 & 23.1 \\
            \hline\hline
            CDP& ours & 36.2& 81.0 & \textbf{92.1} & 37.5   \\
            CDP+DHSM& ours &\textbf{38.0}&\textbf{82.3}& 91.7&\textbf{38.4}  \\
            \Xhline{2\arrayrulewidth}
        \end{tabular}
    \end{center}
\end{table}

\begin{table}\footnotesize
	\caption{Comparison of our proposed method and the state-of-the-art methods on RegDB~\cite{RegDB} dataset.}
	\label{tab:SOTA-RegDB}
	\begin{center}
		\begin{tabular}{c|c|c c c|c}
			\Xhline{2\arrayrulewidth}
			\multicolumn{6}{c}{Visible to Thermal}\\
			\hline
			Method & Reference & r=1 & r=10 & r=20 & mAP\\
			\hline
			Zero-Padding~\cite{SYSU-MM01,BDTR}&ICCV 2017 & 17.8 & 34.2 & 44.4 & 18.9 \\
			TONE~\cite{HCML,BDTR}        &AAAI 2018 & 16.9 & 34.0 & 44.1 & 14.9 \\
			HCML~\cite{HCML}             &AAAI 2018 & 24.4 & 47.5 & 56.8 & 20.1 \\
			BDTR~\cite{BDTR}             &IJCAI 2018& 33.5 & 58.4 & 67.5 & 31.8 \\
			$\text{D}^2\text{RL}$~\cite{D2RL} &CVPR 2019 & 43.4 & 66.1 & 76.3 & 44.1 \\
			HSME~\cite{D-HSME}           &AAAI 2019 & 41.3 & 65.2 & 75.1 & 38.3 \\
			D-HSME~\cite{D-HSME}         &AAAI 2019 & 50.9 & 73.4 & 81.7 & 47.0 \\
			\hline\hline
			CDP& ours &63.8& 83.4 &\textbf{90.0} & 61.9   \\
			CDP+DHSM& ours &\textbf{65.0}&\textbf{83.5}&89.6&\textbf{62.7}  \\
			\Xhline{2\arrayrulewidth}
			\noalign{\vskip 0.5em}    
			\Xhline{2\arrayrulewidth}
			\multicolumn{6}{c}{Thermal to Visible}\\
			\hline
			Method & Reference & r=1 & r=10 & r=20 & mAP\\
			\hline
			Zero-Padding~\cite{SYSU-MM01,BDTR}&ICCV 2017 & 16.6 & 34.7 & 44.3 & 17.8 \\
			TONE~\cite{HCML,BDTR}        &AAAI 2018 & 13.9 & 30.1 & 40.1 & 17.0 \\
			HCML~\cite{HCML}             &AAAI 2018 & 21.7 & 45.0 & 55.6 & 22.2 \\
			BDTR~\cite{BDTR}             &IJCAI 2018& 32.7 & 58.0 & 68.9 & 31.1 \\
			HSME~\cite{D-HSME}           &AAAI 2019 & 40.7 & 65.4 & 75.3 & 37.5 \\
			D-HSME~\cite{D-HSME}         &AAAI 2019 & 50.2 & 72.4 & 81.1 & 46.2 \\
			\hline\hline
			CDP& ours & 64.4 & \textbf{84.5} & 90.8 & 61.5   \\
			CDP+DHSM& ours &\textbf{65.3}&\textbf{84.5}&\textbf{91.0}&\textbf{62.1}  \\
			\Xhline{2\arrayrulewidth}
		\end{tabular}
	\end{center}
\end{table}

As shown in Table~\ref{tab:SOTA-SYSU}, our CDP model achieves a rank-1 accuracy of $36.2\%$, which is $7.3\%$ absolute points higher than the recent $\text{D}^2\text{RL}$~\cite{D2RL} work. We also achieve an mAP of $37.5\%$, outperforming the state-of-the-art methods by a large margin. With the help of DHSM, the performance could be further improved to a rank-1 accuracy of $38.0\%$ and an mAP score of $38.4\%$.

As shown in Table~\ref{tab:SOTA-RegDB}, We achieve the best performance in both visible-to-thermal and thermal-to-visible settings, which has demonstrated the efficiency of our proposed method. Note that, on RegDB dataset, rank-1 accuracy is improved from
$50.9\%$ by D-HSME~\cite{D-HSME} to $65.0\%$, and mAP is improved from $47.0\%$ to $62.7\%$. To the best of our knowledge, it is the highest score ever reported, setting up a new state-of-the-art.

The deep zero-padding~\cite{SYSU-MM01} method only utilizes a sub-network for each modality, which limits the capacity of the network and leads to inferior performance. After this pioneering work, HCML~\cite{HCML} is proposed, which jointly
optimizes the modality-specific and modality-shared metrics. But the two-stage training process requires human intervention, limiting its wide large-scale application in real-world scenarios. While our CDP model can be trained in an end-to-end way without interruption.

BDTR~\cite{BDTR} proposes a dual-path network with a novel bi-directional dual-constrained top-ranking loss. We also leverage the power of ranking loss, adding a batch hard triplet loss. But we use a one-stream network and carefully design the way of training data generation. With the help of the cross-spectrum image generation method and dual-subspace pairing strategy, we achieve better performance.

HSME~\cite{D-HSME} maps the input images onto a hypersphere manifold, and acquires decorrelated features with a two-stage training scheme. While our proposed model can be trained in an end-to-end way, and dynamic hard spectrum mining strategy can also be incorporated into the training process with no need for multi-stage training.

GAN-base methods like cmGAN~\cite{cmGAN} and $\text{D}^2\text{RL}$~\cite{D2RL} are also proposed in recently years. Different from those methods, our cross-image generation method does not need to train an extra GAN model, and generating new images is simple, with high computational efficiency. Unlike $\text{D}^2\text{RL}$~\cite{D2RL}, we do not require extra Market-1501~\cite{Market-1501} dataset for pre-training.

In conclusion, our proposed method is easy to implement, and can be trained in an end-to-end way requiring no extra data, achieving a new state-of-the-art performance.

\subsection{Ablation Study}

\textbf{$\boldsymbol{PK}$ Sampling and Loss Function.} To investigate the influence of $PK$ sampling strategy and different loss function choices, we design a series of baseline models based on our one-stream network without cross-spectrum image generation method or dual-subspace pairing strategy. As shown in Table~\ref{tab:PK}, baseline-1 without $PK$ sampling obtains an inferior performance. After adopting the $PK$ sampling strategy, baseline-2 achieves much better performance, as $PK$ sampling can balance the instance numbers among different persons. Baseline-3 with only batch hard triplet loss gets an inferior performance, which demonstrates the necessity of classification loss (identity loss). Combining $PK$ sampling, classification loss and batch hard triplet loss, baseline-4 achieves the best performance.

\begin{table}\footnotesize
    \caption{Influence of $PK$ sampling strategy and loss function for baseline network on SYSU-MM01~\cite{SYSU-MM01}.}
    \label{tab:PK}
    \begin{center}
        \begin{tabular}{c|c|c|c|c c c|c}
			\Xhline{2\arrayrulewidth}
            Method & $PK$? & $L_{cls}$ & $L_{tri}$ & r=1 & r=10 & r=20 & mAP\\
            \hline
            baseline-1  &\xmark&\cmark&\xmark&16.6&56.1&72.2&18.6\\
            baseline-2  &\cmark&\cmark&\xmark&25.3&71.9&84.5&27.2\\
            baseline-3  &\cmark&\xmark&\cmark&14.1&59.3&76.8&19.3\\
            baseline-4&\cmark&\cmark&\cmark&\textbf{29.1}&\textbf{74.9}&\textbf{87.7}&\textbf{31.2}\\
			\Xhline{2\arrayrulewidth}
        \end{tabular}
    \end{center}
\end{table}

\textbf{Dual-Subspace Pairing.} The influence of the dual-subspace pairing strategy is shown in Table~\ref{tab:Dual-subspace}. After adding the proposed dual-subspace pairing strategy, our CDP-1 model outperforms baseline-4 with only classification loss, which shows the efficiency of our proposed dual-subspace pairing strategy. With the help of batch hard triplet loss, the performance is further improved and the resulting CDP model achieves a rank-1 accuracy of $36.2\%$.

\begin{table}\footnotesize
    \caption{Influence of dual-subspace pairing strategy on SYSU-MM01~\cite{SYSU-MM01}.}
    \label{tab:Dual-subspace}
    \begin{center}
        \begin{tabular}{c|c|c|c c c|c}
			\Xhline{2\arrayrulewidth}
            Method & $L_{cls}$ & $L_{tri}$ & r=1 & r=10 & r=20 & mAP\\
            \hline
            baseline-4 &\cmark&\cmark&29.1&74.9&87.7&31.2\\
            CDP-1      &\cmark&\xmark&32.2&80.2&90.9&33.3\\
            CDP&\cmark&\cmark&\textbf{36.2}&\textbf{81.0}&\textbf{92.1}&\textbf{37.5}\\
			\Xhline{2\arrayrulewidth}
        \end{tabular}
    \end{center}
\end{table}

\textbf{Cross-Spectrum Image Generation.} For each original image, we generate a paired image randomly chosen from four modalities, which are red, green, blue, and gray spectrum respectively. To further investigate the influence of the cross-spectrum image generation method, we reduce the number of generated modalities to one for CDP-2 and three for CDP-3 model. The result is shown in Table~\ref{tab:Cross-spectrum}. For the CDP-2 model, we only generate the gray spectrum counterpart for each original image, while for the CDP-3 model, we obtain the paired image through randomly choosing from red, green, and blue channels. As we can see from Table~\ref{tab:Cross-spectrum}, with the increase of modality number, the performance improves. And cross-spectrum image generation with all RGB and gray spectrum achieves the best performance, which conforms to our motivation that after seeing input images from more modalities, the network is guided to discover more discriminative cross-spectrum shared features across all modalities.

\begin{table}\footnotesize
	\caption{Influence of reducing color channel on SYSU-MM01~\cite{SYSU-MM01}. The more spectrum we use, the better performance we get.}
	\label{tab:rgb-channel}
	\begin{center}
		\begin{tabular}{c|c|c|c|c|c c c|c}
			\Xhline{2\arrayrulewidth}
			Gray? & R? & G? & B? & \#modalities & r=1 & r=10 & r=20 & mAP\\
			\hline
			\xmark &\xmark &\xmark &\xmark &0&29.1&74.9&87.7&31.2\\
			\cmark &\xmark &\xmark &\xmark &1&33.2&80.1&90.9&35.2\\
			\cmark &\cmark &\xmark &\xmark &2&33.5&81.5&91.9&35.7\\
			\cmark &\xmark &\cmark &\xmark &2&33.9&82.0&92.4&36.3\\
			\cmark &\xmark &\xmark &\cmark &2&35.2&81.2&91.7&36.5\\
			\cmark &\xmark &\cmark &\cmark &3&34.8&80.5&91.6&36.8\\
			\cmark &\cmark &\xmark &\cmark &3&35.8&80.8&91.4&37.2\\
			\cmark &\cmark &\cmark &\xmark &3&35.4&\textbf{82.4}&\textbf{92.1}&37.0\\
			\cmark &\cmark &\cmark &\cmark &4&\textbf{36.2}&81.0&\textbf{92.1}&\textbf{37.5}\\
			\Xhline{2\arrayrulewidth}
		\end{tabular}
	\end{center}
\end{table}

\begin{figure}[ht]
	\begin{center}
		\includegraphics[width=0.9\linewidth]{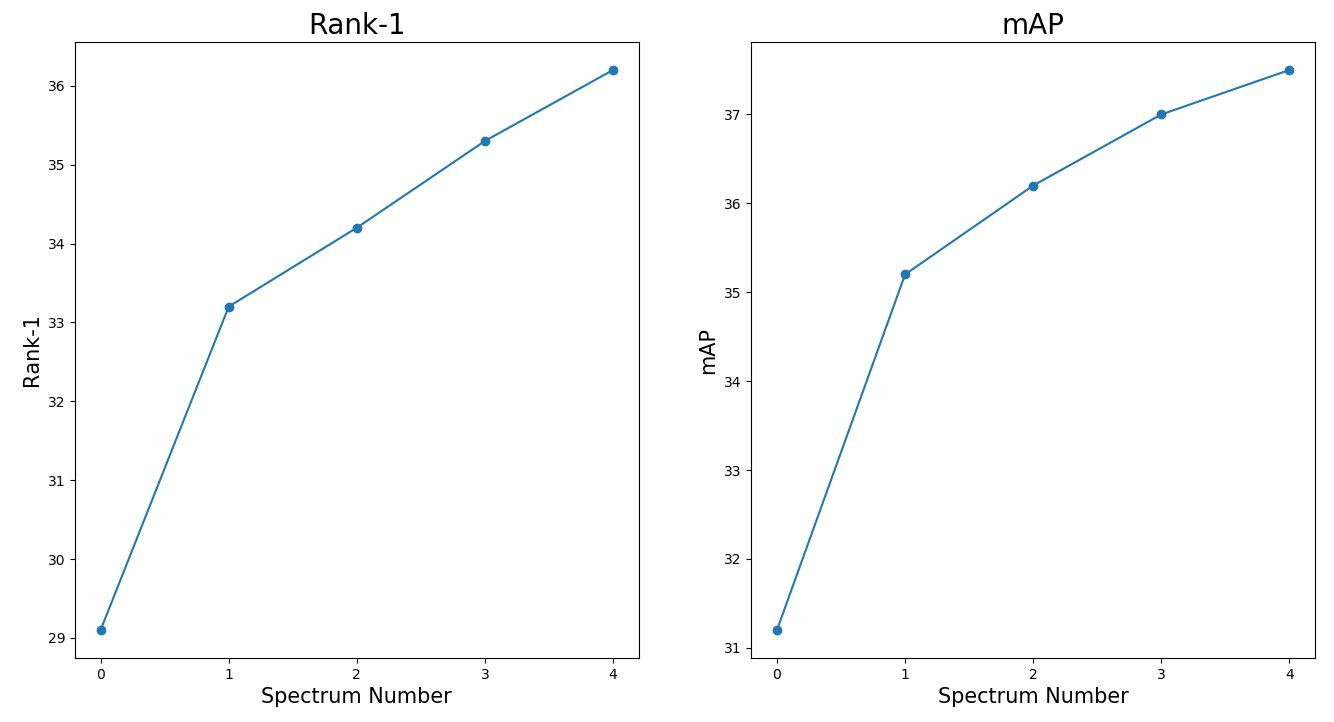}
	\end{center}
	\caption{Influence of spectrum number on SYSU-MM01~\cite{SYSU-MM01}. The scores are taken from Table~\ref{tab:rgb-channel}, and we take the average of three experiments when the modality number is 2 or 3. As we can see, with more modalities, the performance becomes better, which has demonstrated the efficiency of our cross-spectrum dual-subspace pairing method.}
	\label{fig:reduce-color}
\end{figure}

\textbf{Reducing RGB channel} Considering the gray spectrum is a linear combination of RGB spectrum, we also conduct an ablation experiment to see how the performance will change if we reduce RGB channel and only use one or two color channels. The result is shown in Table~\ref{tab:rgb-channel}. As we can see, without any cross-spectrum image generation, the result in the first row is inferior. As we increase the spectrum number, the performance improves, which can be seen in Fig.~\ref{fig:reduce-color}. Since we adopt the dual-subspace pairing strategy, each original input image is paired with a generated image regardless of which spectrum it is sampled from, therefor training with only gray spectrum, gray spectrum with a single color spectrum (red, green, or blue), or gray spectrum with all color spectrum has exactly the same computation cost. The more spectrum we use, the better performance we get, so we use all spectrum in our final CDP model. This is also the reason why we choose a three channel model input, R-G-B and R-R-R are different spectrum, and we want the model to see more input spectrum to extract cross-spectrum discriminated features. In addition, using three-channel input only adds computation cost in the first layer, which is minor, and the pre-trained model we use also has three-channel input.

\textbf{Different Color Spaces} In addition to RGB color space, a question naturally arises in the mind, can we use other color spaces like YUV or HSV instead of RGB color space and maybe they can achieve a better performance? or can we combine RGB color space as well as other color spaces like YUV or HSV to further promote the performance? So we also conduct a ablation study replacing or combing RGB color space with YUV and HSV color spaces, and the result is that all these experiments end up with no convergence and fail to obtain a usable model. After examining the training data and model settings, we attribute it to three reasons: (1) the YUV and HSV images are too visually different with RGB images, and (2) the pre-trained model on ImageNet we use for model initialization is trained on RGB images, so it is not directly applicable for YUV and HSV images, and (3) the key point of this paper is cross-spectrum dual subspace pairing, and RGB images have homogeneous channels which can be equally sampled and replaced, while YUV and HSV images have heterogeneous channels encoding different attributes (for example, HSV channels represent hue, saturation and value respectively), which is not suitable for our model.

\begin{table}\footnotesize
    \caption{Influence of cross-spectrum image generation method on SYSU-MM01~\cite{SYSU-MM01}. With the increase of modality numbers, the performance improves.}
    \label{tab:Cross-spectrum}
    \begin{center}
        \begin{tabular}{c|c|c|c|c c c|c}
			\Xhline{2\arrayrulewidth}
            Method & RGB? & Gray? & \#modalities & r=1 & r=10 & r=20 & mAP\\
            \hline
            CDP-2      &\xmark&\cmark&1&33.2&80.1&90.9&35.2\\
            CDP-3      &\cmark&\xmark&3&34.9&\textbf{81.6}&92.0&36.9\\
            CDP&\cmark&\cmark&4&\textbf{36.2}&81.0&\textbf{92.1}&\textbf{37.5}\\
			\Xhline{2\arrayrulewidth}
        \end{tabular}
    \end{center}
\end{table}

\begin{figure}[h]
    \begin{center}
        \includegraphics[width=0.95\linewidth]{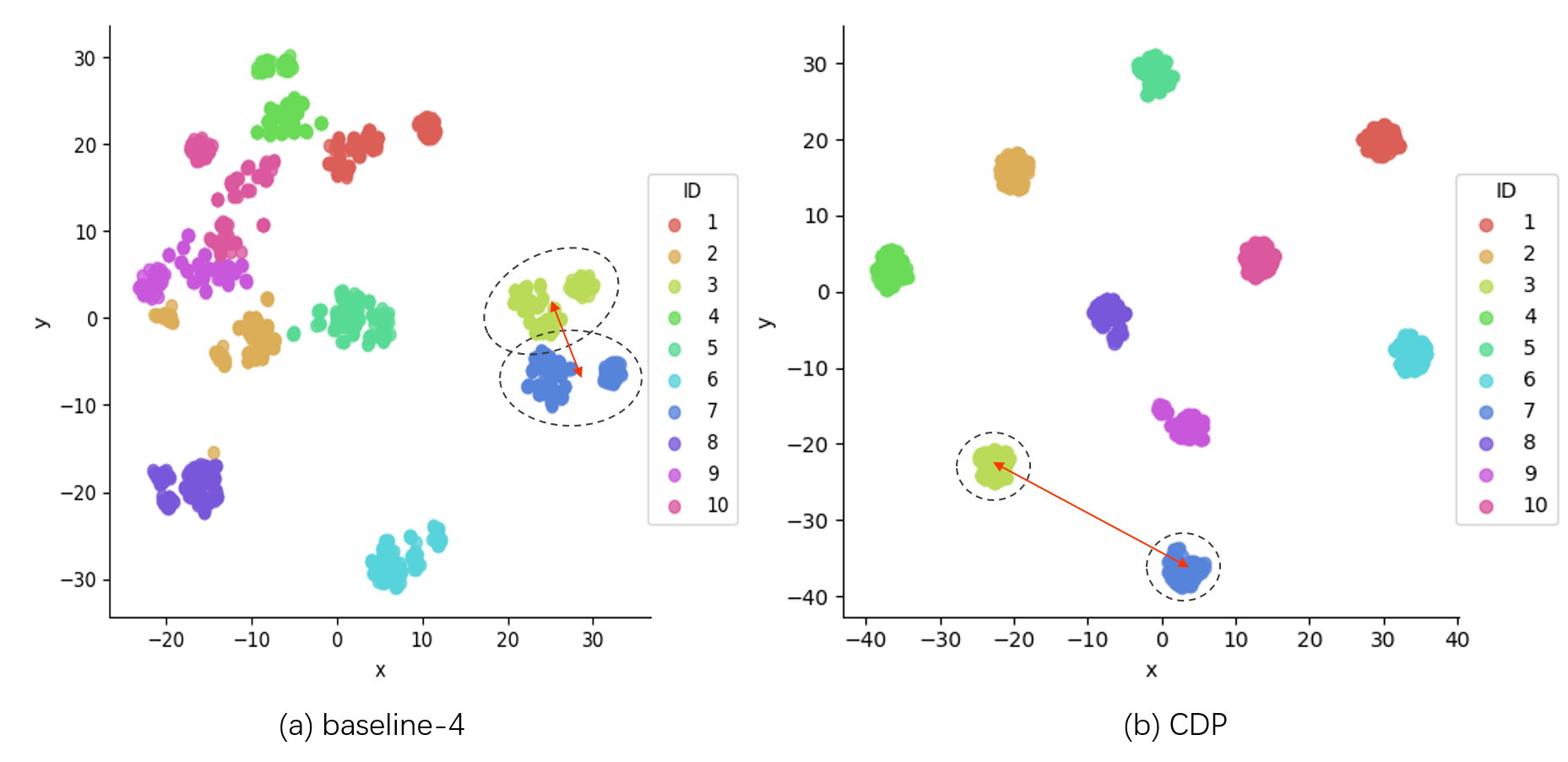}
    \end{center}
    \caption{Visualization of feature embedding of 10 persons from SYSU-MM01~\cite{SYSU-MM01} dataset using t-SNE~\cite{t-SNE}. (a) the feature distribution of baseline-4, and (b) the feature distribution of CDP. We can observe that baseline-4 has two modality distribution centers, while CDP maps input images of different modalities into an unified feature distribution. The intra-person distance is reduced (dotted circle), while the inter-person distance is increased (red line).}
    \label{fig:vis_feat}
\end{figure}

\textbf{Visualization of Feature Embedding.} To investigate the distribution of different person in the embedding space, we randomly choose 10 persons and extract the features of RGB images, infrared images and generated images of each person. We use a t-SNE~\cite{t-SNE} method to visualize the features in 2-dimension space. As shown in Fig.~\ref{fig:vis_feat}, for each person, the baseline-4 model has two distribution corresponding to two different modalities, while the CDP maps input images of different modalities into a unified feature distribution, for example, the blue dots of the person with ID 7. Note that baseline-4 and CDP models both adopt a $PK$ sampling strategy with the same loss function, and the difference is that CDP adds a cross-spectrum dual-space pairing strategy. Our motivation is that a person should have a fixed underlying appearance, although appearing different in different modalities. If we want to do cross-modalities searching, we should focus on the cross-spectrum shared features. As shown in Fig.~\ref{fig:vis_feat}, our proposed method can map images of different modalities into a shared cross-spectrum space, and achieves a new state-of-the-art performance.

\section{Conclusion}
\label{Conclusion}

In this paper, we propose a cross-spectrum image generation method to produce multi-spectrum images, which can enrich the diversity of modalities in training process. After seeing more training modalities, the network can do better to discover more discriminative shared features across all modalities. Both appearance discrepancy and modality discrepancy exist in RGB-infrared ReID task, and intra-person distance is often larger than the inter-person distance due to the large modality discrepancy. To address this problem, we propose a dual-subspace pairing strategy which pairs each original image with a generated image with the same underlying appearance in a different spectrum, therefore, the network can focus on eliminating the modality discrepancy without the interference of appearance discrepancy. To evaluate the performance of our proposed method, we design a one-stream convolutional neural network for RGB-infrared cross-modality ReID task. Furthermore, we also propose a dynamic hard spectrum mining method to generate more hard samples during the training process, based on current model state. Experimental results on two public datasets, SYSU-MM01~\cite{SYSU-MM01} with RGB + near-infrared images and RegDB~\cite{RegDB} with RGB + far-infrared images, have demonstrated the efficiency and generality of our proposed method, which outperform the current state-of-the-art approaches by a large margin.

\section*{ACKNOWLEDGMENT}
This work is supported by the National Natural Science Foundation of China (No. 61633019) and the Science Foundation of Chinese Aerospace Industry (JCKY2018204B053).



%

\small{
    \bibliographystyle{IEEEtran}
    \bibliography{egbib}

\begin{thebibliography}{10}
\providecommand{\url}[1]{#1}
\csname url@samestyle\endcsname
\providecommand{\newblock}{\relax}
\providecommand{\bibinfo}[2]{#2}
\providecommand{\BIBentrySTDinterwordspacing}{\spaceskip=0pt\relax}
\providecommand{\BIBentryALTinterwordstretchfactor}{4}
\providecommand{\BIBentryALTinterwordspacing}{\spaceskip=\fontdimen2\font plus
\BIBentryALTinterwordstretchfactor\fontdimen3\font minus
  \fontdimen4\font\relax}
\providecommand{\BIBforeignlanguage}[2]{{%
\expandafter\ifx\csname l@#1\endcsname\relax
\typeout{** WARNING: IEEEtran.bst: No hyphenation pattern has been}%
\typeout{** loaded for the language `#1'. Using the pattern for}%
\typeout{** the default language instead.}%
\else
\language=\csname l@#1\endcsname
\fi
#2}}
\providecommand{\BIBdecl}{\relax}
\BIBdecl

\bibitem{DDDM}
Z.~Wang, R.~Hu, C.~Liang, Y.~Yu, J.~Jiang, M.~Ye, J.~Chen, and Q.~Leng,
  ``Zero-{{Shot Person Re}}-identification via {{Cross}}-{{View
  Consistency}},'' \emph{IEEE Transactions on Multimedia}, 2016.

\bibitem{SYSU-MM01}
A.~Wu, W.-S. Zheng, H.-X. Yu, S.~Gong, and J.~Lai, ``{{RGB}}-{{Infrared
  Cross}}-{{Modality Person Re}}-{{Identification}},'' in \emph{The {{IEEE
  International Conference}} on {{Computer Vision}} ({{ICCV}})}, 2017.

\bibitem{WhereandWhen}
L.~Wu, Y.~Wang, J.~Gao, and X.~Li, ``Where-and-{{When}} to {{Look}}: {{Deep
  Siamese Attention Networks}} for {{Video}}-{{Based Person
  Re}}-{{Identification}},'' \emph{IEEE Transactions on Multimedia}, 2019.

\bibitem{yang2017person}
X.~Yang, M.~Wang, and D.~Tao, ``Person re-identification with metric learning
  using privileged information,'' \emph{IEEE Transactions on Image Processing},
  vol.~27, no.~2, pp. 791--805, 2017.

\bibitem{yang2018person}
X.~Yang, P.~Zhou, and M.~Wang, ``Person reidentification via structural deep
  metric learning,'' \emph{IEEE Transactions on Neural Networks and Learning
  Systems}, vol.~30, no.~10, pp. 2987--2998, 2019.

\bibitem{S2S}
S.~Zhou, J.~Wang, R.~Shi, Q.~Hou, Y.~Gong, and N.~Zheng, ``Large {{Margin
  Learning}} in {{Set}}-to-{{Set Similarity Comparison}} for {{Person
  Reidentification}},'' \emph{IEEE Transactions on Multimedia}, 2018.

\bibitem{GLAD_TMM}
L.~Wei, S.~Zhang, H.~Yao, W.~Gao, and Q.~Tian, ``{{GLAD}}:
  {{Global}}\textendash{{Local}}-{{Alignment Descriptor}} for {{Scalable Person
  Re}}-{{Identification}},'' \emph{IEEE Transactions on Multimedia}, 2019.

\bibitem{Triplet_Hard_Loss}
A.~Hermans, L.~Beyer, and B.~Leibe, ``In {{Defense}} of the {{Triplet Loss}}
  for {{Person Re}}-{{Identification}},'' \emph{arXiv 1703.07737}, 2017.

\bibitem{PCB+RPP}
Y.~Sun, L.~Zheng, Y.~Yang, Q.~Tian, and S.~Wang, ``Beyond {{Part Models}}:
  {{Person Retrieval}} with {{Refined Part Pooling}},'' in \emph{The {{European
  Conference}} on {{Computer Vision}} ({{ECCV}})}, 2018.

\bibitem{BDTR}
M.~Ye, Z.~Wang, X.~Lan, and P.~C. Yuen, ``Visible {{Thermal Person
  Re}}-{{Identification}} via {{Dual}}-{{Constrained Top}}-{{Ranking}}.'' in
  \emph{{{IJCAI}}}, 2018.

\bibitem{HCML}
M.~Ye, X.~Lan, J.~Li, and P.~C. Yuen, ``Hierarchical discriminative learning
  for visible thermal person re-identification,'' in \emph{{{AAAI}}}, 2018.

\bibitem{RegDB}
D.~Nguyen, H.~Hong, K.~Kim, and K.~Park, ``Person recognition system based on a
  combination of body images from visible light and thermal cameras,'' vol.~17,
  no.~3, p. 605.

\bibitem{GAN}
I.~Goodfellow, J.~{Pouget-Abadie}, M.~Mirza, B.~Xu, D.~{Warde-Farley},
  S.~Ozair, A.~Courville, and Y.~Bengio, ``Generative {{Adversarial Nets}},''
  in \emph{Advances in {{Neural Information Processing Systems}}}, 2014.

\bibitem{cmGAN}
P.~Dai, R.~Ji, H.~Wang, Q.~Wu, and Y.~Huang, ``Cross-{{Modality Person
  Re}}-{{Identification}} with {{Generative Adversarial Training}}.'' in
  \emph{{{IJCAI}}}, 2018.

\bibitem{D2RL}
Z.~Wang, Z.~Wang, Y.~Zheng, Y.-Y. Chuang, and S.~Satoh, ``Learning to {{Reduce
  Dual}}-level {{Discrepancy}} for {{Infrared}}-{{Visible Person
  Re}}-identification,'' in \emph{The {{IEEE Conference}} on {{Computer
  Vision}} and {{Pattern Recognition}} ({{CVPR}})}, 2019.

\bibitem{LADF}
Z.~Li, S.~Chang, F.~Liang, T.~S. Huang, L.~Cao, and J.~R. Smith, ``Learning
  locally-adaptive decision functions for person verification,'' in \emph{The
  {{IEEE Conference}} on {{Computer Vision}} and {{Pattern Recognition}}
  ({{CVPR}})}, 2013.

\bibitem{LOMO_XQDA}
S.~Liao, Y.~Hu, X.~Zhu, and S.~Z. Li, ``Person re-identification by {{Local
  Maximal Occurrence}} representation and metric learning,'' in \emph{The
  {{IEEE Conference}} on {{Computer Vision}} and {{Pattern Recognition}}
  ({{CVPR}})}, 2015.

\bibitem{KISSME}
M.~Koestinger, M.~Hirzer, P.~Wohlhart, P.~M. Roth, and H.~Bischof, ``Large
  scale metric learning from equivalence constraints,'' in \emph{The {{IEEE
  Conference}} on {{Computer Vision}} and {{Pattern Recognition}} ({{CVPR}})},
  2012.

\bibitem{7451215}
D.~Tao, Y.~Guo, M.~Song, Y.~Li, Z.~Yu, and Y.~Y. Tang, ``Person
  {{Re}}-{{Identification}} by {{Dual}}-{{Regularized KISS Metric Learning}},''
  \emph{IEEE Transactions on Image Processing}, 2016.

\bibitem{LFDA}
C.~Shen, Z.~Jin, Y.~Zhao, Z.~Fu, R.~Jiang, Y.~Chen, and X.-S. Hua, ``Deep
  siamese network with multi-level similarity perception for person
  re-identification,'' in \emph{Proceedings of the 25th {{ACM}} International
  Conference on {{Multimedia}}}, 2017.

\bibitem{k-LFDA}
F.~Xiong, M.~Gou, O.~Camps, and M.~Sznaier, ``Person re-identification using
  kernel-based metric learning methods,'' in \emph{European Conference on
  Computer Vision}, 2014.

\bibitem{Siamese_CNN}
D.~Yi, Z.~Lei, S.~Liao, and S.~Z. Li, ``Deep metric learning for person
  re-identification,'' in \emph{Proceedings of {{International Conference}} on
  {{Pattern Recognition}} ({{ICPR}})}, 2014.

\bibitem{Gated_SCNN}
R.~R. Varior, M.~Haloi, and G.~Wang, ``Gated siamese convolutional neural
  network architecture for human re-identification,'' in \emph{European
  {{Conference}} on {{Computer Vision}} ({{ECCV}})}, 2016.

\bibitem{Contrastive_Loss}
R.~Hadsell, S.~Chopra, and Y.~LeCun, ``Dimensionality reduction by learning an
  invariant mapping,'' in \emph{The {{IEEE Conference}} on {{Computer Vision}}
  and {{Pattern Recognition}} ({{CVPR}})}, vol.~2, 2006.

\bibitem{Triplet_Loss}
K.~Q. Weinberger and L.~K. Saul, ``Distance metric learning for large margin
  nearest neighbor classification,'' \emph{The Journal of Machine Learning
  Research (JMLR)}, 2009.

\bibitem{FaceNet}
F.~Schroff, D.~Kalenichenko, and J.~Philbin, ``Facenet: {{A}} unified embedding
  for face recognition and clustering,'' in \emph{The {{IEEE Conference}} on
  {{Computer Vision}} and {{Pattern Recognition}} ({{CVPR}})}, 2015.

\bibitem{dingDeepFeatureLearning2015}
S.~Ding, L.~Lin, G.~Wang, and H.~Chao, ``Deep feature learning with relative
  distance comparison for person re-identification,'' \emph{Pattern
  Recognition}, 2015.

\bibitem{Improved_Triplet_Loss}
D.~Cheng, Y.~Gong, S.~Zhou, J.~Wang, and N.~Zheng, ``Person
  {{Re}}-{{Identification}} by {{Multi}}-{{Channel Parts}}-{{Based CNN With
  Improved Triplet Loss Function}},'' in \emph{The {{IEEE Conference}} on
  {{Computer Vision}} and {{Pattern Recognition}} ({{CVPR}})}, 2016.

\bibitem{IDE}
L.~Zheng, Y.~Yang, and A.~G. Hauptmann, ``Person {{Re}}-identification:
  {{Past}}, {{Present}} and {{Future}},'' \emph{arXiv 1610.02984}, 2016.

\bibitem{chen2017multi}
W.~Chen, X.~Chen, J.~Zhang, and K.~Huang, ``A multi-task deep network for
  person re-identification,'' in \emph{Thirty-{{First AAAI Conference}} on
  {{Artificial Intelligence}}}, 2017.

\bibitem{Quad_Loss}
------, ``Beyond {{Triplet Loss}}: {{A Deep Quadruplet Network}} for {{Person
  Re}}-{{Identification}},'' in \emph{The {{IEEE Conference}} on {{Computer
  Vision}} and {{Pattern Recognition}} ({{CVPR}})}, 2017.

\bibitem{Multi-index}
Z.~Zhang, Y.~Xie, W.~Zhang, and Q.~Tian, ``Effective {{Image Retrieval}} via
  {{Multilinear Multi}}-index {{Fusion}},'' \emph{IEEE Transactions on
  Multimedia}, 2019.

\bibitem{Re-ranking}
Z.~Zhong, L.~Zheng, D.~Cao, and S.~Li, ``Re-ranking {{Person
  Re}}-identification with k-reciprocal {{Encoding}},'' in \emph{The {{IEEE
  Conference}} on {{Computer Vision}} and {{Pattern Recognition}} ({{CVPR}})},
  2017.

\bibitem{Ranking-agg}
M.~Ye, C.~Liang, Y.~Yu, Z.~Wang, Q.~Leng, C.~Xiao, J.~Chen, and R.~Hu, ``Person
  {{Reidentification}} via {{Ranking Aggregation}} of {{Similarity Pulling}}
  and {{Dissimilarity Pushing}},'' \emph{IEEE Transactions on Multimedia},
  2016.

\bibitem{RO-re-ranking}
Z.~Wang, J.~Jiang, Y.~Yu, and S.~Satoh, ``Incremental {{Re}}-identification by
  {{Cross}}-{{Direction}} and {{Cross}}-{{Ranking Adaption}},'' \emph{IEEE
  Transactions on Multimedia}, 2019.

\bibitem{PDC}
C.~Su, J.~Li, S.~Zhang, J.~Xing, W.~Gao, and Q.~Tian, ``Pose-driven {{Deep
  Convolutional Model}} for {{Person Re}}-identification,'' in \emph{{{ICCV}}},
  2017.

\bibitem{SpindleNet}
H.~Zhao, M.~Tian, S.~Sun, J.~Shao, J.~Yan, S.~Yi, X.~Wang, and X.~Tang,
  ``Spindle {{Net}}: {{Person Re}}-{{Identification With Human Body Region
  Guided Feature Decomposition}} and {{Fusion}},'' in \emph{The {{IEEE
  Conference}} on {{Computer Vision}} and {{Pattern Recognition}} ({{CVPR}})},
  2017.

\bibitem{PoseBox}
L.~Zheng, Y.~Huang, H.~Lu, and Y.~Yang, ``Pose {{Invariant Embedding}} for
  {{Deep Person Re}}-identification,'' \emph{IEEE Transactions on Image
  Processing}, 2019.

\bibitem{PN-GAN}
X.~Qian, Y.~Fu, T.~Xiang, W.~Wang, J.~Qiu, Y.~Wu, Y.-G. Jiang, and X.~Xue,
  ``Pose-{{Normalized Image Generation}} for {{Person Re}}-identification,'' in
  \emph{The {{European Conference}} on {{Computer Vision}} ({{ECCV}})}, 2018.

\bibitem{FD-GAN}
Y.~Ge, Z.~Li, H.~Zhao, G.~Yin, S.~Yi, X.~Wang, and H.~Li, ``{{FD}}-{{GAN}}:
  {{Pose}}-guided {{Feature Distilling GAN}} for {{Robust Person
  Re}}-identification,'' in \emph{{{NIPS}}}, 2018.

\bibitem{Siamese-LSTM}
R.~R. Varior, B.~Shuai, J.~Lu, D.~Xu, and G.~Wang, ``A siamese long short-term
  memory architecture for human re-identification,'' in \emph{European
  {{Conference}} on {{Computer Vision}} ({{ECCV}})}, 2016.

\bibitem{latent_parts}
D.~Li, X.~Chen, Z.~Zhang, and K.~Huang, ``Learning {{Deep Context}}-{{Aware
  Features Over Body}} and {{Latent Parts}} for {{Person
  Re}}-{{Identification}},'' in \emph{The {{IEEE Conference}} on {{Computer
  Vision}} and {{Pattern Recognition}} ({{CVPR}})}, 2017.

\bibitem{STN}
M.~Jaderberg, K.~Simonyan, A.~Zisserman, and {Kavukcuoglu, Koray}, ``Spatial
  transformer networks,'' in \emph{Advances in {{Neural Information Processing
  Systems}}}, 2015.

\bibitem{cluster_parts}
H.~Yao, S.~Zhang, Y.~Zhang, J.~Li, and Q.~Tian, ``Deep {{Representation
  Learning}} with {{Part Loss}} for {{Person Re}}-{{Identification}},''
  \emph{IEEE Transactions on Image Processing}, 2019.

\bibitem{matsukawaPersonReidentificationUsing2016}
T.~Matsukawa and E.~Suzuki, ``Person re-identification using {{CNN}} features
  learned from combination of attributes,'' in \emph{Proceedings of
  {{International Conference}} on {{Pattern Recognition}} ({{ICPR}})}, 2016.

\bibitem{xiaoLearningDeepFeature2016}
T.~Xiao, H.~Li, W.~Ouyang, and X.~Wang, ``Learning {{Deep Feature
  Representations With Domain Guided Dropout}} for {{Person
  Re}}-{{Identification}},'' in \emph{The {{IEEE Conference}} on {{Computer
  Vision}} and {{Pattern Recognition}} ({{CVPR}})}, 2016.

\bibitem{Mask-Guided}
C.~Song, Y.~Huang, W.~Ouyang, and L.~Wang, ``Mask-{{Guided Contrastive
  Attention Model}} for {{Person Re}}-{{Identification}},'' in \emph{The {{IEEE
  Conference}} on {{Computer Vision}} and {{Pattern Recognition}} ({{CVPR}})},
  2018.

\bibitem{Mutual}
Y.~Zhang, T.~Xiang, T.~M. Hospedales, and H.~Lu, ``Deep {{Mutual Learning}},''
  in \emph{The {{IEEE Conference}} on {{Computer Vision}} and {{Pattern
  Recognition}} ({{CVPR}})}, 2018.

\bibitem{SGGNN}
Y.~Shen, H.~Li, S.~Yi, D.~Chen, and X.~Wang, ``Person {{Re}}-identification
  with {{Deep Similarity}}-{{Guided Graph Neural Network}},'' in \emph{The
  {{European Conference}} on {{Computer Vision}} ({{ECCV}})}, 2018.

\bibitem{FAPL}
G.~Ding, S.~Zhang, S.~Khan, Z.~Tang, J.~Zhang, and F.~Porikli, ``Feature
  {{Affinity}} based {{Pseudo Labeling}} for {{Semi}}-supervised {{Person
  Re}}-identification,'' \emph{IEEE Transactions on Multimedia}, 2019.

\bibitem{Context-Aware-Hypergraph}
S.~Sunderrajan and B.~S. Manjunath, ``Context-{{Aware Hypergraph Modeling}} for
  {{Re}}-identification and {{Summarization}},'' \emph{IEEE Transactions on
  Multimedia}, 2016.

\bibitem{Market-1501}
L.~Zheng, L.~Shen, L.~Tian, S.~Wang, J.~Wang, and Q.~Tian, ``Scalable {{Person
  Re}}-{{Identification}}: {{A Benchmark}},'' in \emph{The {{IEEE International
  Conference}} on {{Computer Vision}} ({{ICCV}})}, 2015.

\bibitem{ResNet}
K.~He, X.~Zhang, S.~Ren, and J.~Sun, ``Deep residual learning for image
  recognition,'' in \emph{The {{IEEE Conference}} on {{Computer Vision}} and
  {{Pattern Recognition}} ({{CVPR}})}, 2016.

\bibitem{sunSVDNetPedestrianRetrieval2017}
Y.~Sun, L.~Zheng, W.~Deng, and S.~Wang, ``{{SVDNet}} for {{Pedestrian
  Retrieval}},'' in \emph{The {{IEEE International Conference}} on {{Computer
  Vision}} ({{ICCV}})}.

\bibitem{D-HSME}
Y.~Hao, N.~Wang, J.~Li, and X.~Gao, ``{{HSME}}: Hypersphere manifold embedding
  for visible thermal person re-identification,'' in \emph{Proceedings of the
  {{AAAI}} Conference on Artificial Intelligence}, vol.~33, pp. 8385--8392.

\bibitem{Adam}
D.~Kingma and J.~Ba, ``Adam: {{A}} method for stochastic optimization,'' in
  \emph{International {{Conference}} on {{Learning Representations}}
  ({{ICLR}})}, 2015.

\bibitem{t-SNE}
L.~van~der Maaten and G.~Hinton, ``Visualizing data using t-{{SNE}},''
  \emph{Journal of machine learning research}, 2008.

\end{thebibliography}


\begin{thebibliography}{10}
\providecommand{\url}[1]{#1}
\csname url@samestyle\endcsname
\providecommand{\newblock}{\relax}
\providecommand{\bibinfo}[2]{#2}
\providecommand{\BIBentrySTDinterwordspacing}{\spaceskip=0pt\relax}
\providecommand{\BIBentryALTinterwordstretchfactor}{4}
\providecommand{\BIBentryALTinterwordspacing}{\spaceskip=\fontdimen2\font plus
\BIBentryALTinterwordstretchfactor\fontdimen3\font minus
  \fontdimen4\font\relax}
\providecommand{\BIBforeignlanguage}[2]{{%
\expandafter\ifx\csname l@#1\endcsname\relax
\typeout{** WARNING: IEEEtran.bst: No hyphenation pattern has been}%
\typeout{** loaded for the language `#1'. Using the pattern for}%
\typeout{** the default language instead.}%
\else
\language=\csname l@#1\endcsname
\fi
#2}}
\providecommand{\BIBdecl}{\relax}
\BIBdecl

\bibitem{SYSU-MM01}
A.~Wu, W.-S. Zheng, H.-X. Yu, S.~Gong, and J.~Lai, ``{{RGB}}-{{Infrared
  Cross}}-{{Modality Person Re}}-{{Identification}},'' in \emph{The {{IEEE
  International Conference}} on {{Computer Vision}} ({{ICCV}})}, 2017.

\bibitem{BDTR}
M.~Ye, Z.~Wang, X.~Lan, and P.~C. Yuen, ``Visible {{Thermal Person
  Re}}-{{Identification}} via {{Dual}}-{{Constrained Top}}-{{Ranking}}.'' in
  \emph{{{IJCAI}}}, 2018.

\bibitem{HCML}
M.~Ye, X.~Lan, J.~Li, and P.~C. Yuen, ``Hierarchical discriminative learning
  for visible thermal person re-identification,'' in \emph{{{AAAI}}}, 2018.

\bibitem{GAN}
I.~Goodfellow, J.~{Pouget-Abadie}, M.~Mirza, B.~Xu, D.~{Warde-Farley},
  S.~Ozair, A.~Courville, and Y.~Bengio, ``Generative {{Adversarial Nets}},''
  in \emph{Advances in {{Neural Information Processing Systems}}}, 2014.

\bibitem{cmGAN}
P.~Dai, R.~Ji, H.~Wang, Q.~Wu, and Y.~Huang, ``Cross-{{Modality Person
  Re}}-{{Identification}} with {{Generative Adversarial Training}}.'' in
  \emph{{{IJCAI}}}, 2018.

\bibitem{D2RL}
Z.~Wang, Z.~Wang, Y.~Zheng, Y.-Y. Chuang, and S.~Satoh, ``Learning to {{Reduce
  Dual}}-level {{Discrepancy}} for {{Infrared}}-{{Visible Person
  Re}}-identification,'' in \emph{The {{IEEE Conference}} on {{Computer
  Vision}} and {{Pattern Recognition}} ({{CVPR}})}, 2019.

\bibitem{Triplet_Hard_Loss}
A.~Hermans, L.~Beyer, and B.~Leibe, ``In {{Defense}} of the {{Triplet Loss}}
  for {{Person Re}}-{{Identification}},'' \emph{arXiv 1703.07737}, 2017.

\bibitem{Market-1501}
L.~Zheng, L.~Shen, L.~Tian, S.~Wang, J.~Wang, and Q.~Tian, ``Scalable {{Person
  Re}}-{{Identification}}: {{A Benchmark}},'' in \emph{The {{IEEE International
  Conference}} on {{Computer Vision}} ({{ICCV}})}, 2015.

\bibitem{LADF}
Z.~Li, S.~Chang, F.~Liang, T.~S. Huang, L.~Cao, and J.~R. Smith, ``Learning
  locally-adaptive decision functions for person verification,'' in \emph{The
  {{IEEE Conference}} on {{Computer Vision}} and {{Pattern Recognition}}
  ({{CVPR}})}, 2013.

\bibitem{LOMO_XQDA}
S.~Liao, Y.~Hu, X.~Zhu, and S.~Z. Li, ``Person re-identification by {{Local
  Maximal Occurrence}} representation and metric learning,'' in \emph{The
  {{IEEE Conference}} on {{Computer Vision}} and {{Pattern Recognition}}
  ({{CVPR}})}, 2015.

\bibitem{LFDA}
C.~Shen, Z.~Jin, Y.~Zhao, Z.~Fu, R.~Jiang, Y.~Chen, and X.-S. Hua, ``Deep
  siamese network with multi-level similarity perception for person
  re-identification,'' in \emph{Proceedings of the 25th {{ACM}} International
  Conference on {{Multimedia}}}, 2017.

\bibitem{k-LFDA}
F.~Xiong, M.~Gou, O.~Camps, and M.~Sznaier, ``Person re-identification using
  kernel-based metric learning methods,'' in \emph{European Conference on
  Computer Vision}, 2014.

\bibitem{Siamese_CNN}
D.~Yi, Z.~Lei, S.~Liao, and S.~Z. Li, ``Deep metric learning for person
  re-identification,'' in \emph{Proceedings of {{International Conference}} on
  {{Pattern Recognition}} ({{ICPR}})}, 2014.

\bibitem{Gated_SCNN}
R.~R. Varior, M.~Haloi, and G.~Wang, ``Gated siamese convolutional neural
  network architecture for human re-identification,'' in \emph{European
  {{Conference}} on {{Computer Vision}} ({{ECCV}})}, 2016.

\bibitem{Contrastive_Loss}
R.~Hadsell, S.~Chopra, and Y.~LeCun, ``Dimensionality reduction by learning an
  invariant mapping,'' in \emph{The {{IEEE Conference}} on {{Computer Vision}}
  and {{Pattern Recognition}} ({{CVPR}})}, vol.~2, 2006.

\bibitem{Triplet_Loss}
K.~Q. Weinberger and L.~K. Saul, ``Distance metric learning for large margin
  nearest neighbor classification,'' \emph{The Journal of Machine Learning
  Research (JMLR)}, 2009.

\bibitem{FaceNet}
F.~Schroff, D.~Kalenichenko, and J.~Philbin, ``Facenet: {{A}} unified embedding
  for face recognition and clustering,'' in \emph{The {{IEEE Conference}} on
  {{Computer Vision}} and {{Pattern Recognition}} ({{CVPR}})}, 2015.

\bibitem{dingDeepFeatureLearning2015}
S.~Ding, L.~Lin, G.~Wang, and H.~Chao, ``Deep feature learning with relative
  distance comparison for person re-identification,'' \emph{Pattern
  Recognition}, 2015.

\bibitem{Improved_Triplet_Loss}
D.~Cheng, Y.~Gong, S.~Zhou, J.~Wang, and N.~Zheng, ``Person
  {{Re}}-{{Identification}} by {{Multi}}-{{Channel Parts}}-{{Based CNN With
  Improved Triplet Loss Function}},'' in \emph{The {{IEEE Conference}} on
  {{Computer Vision}} and {{Pattern Recognition}} ({{CVPR}})}, 2016.

\bibitem{IDE}
L.~Zheng, Y.~Yang, and A.~G. Hauptmann, ``Person {{Re}}-identification:
  {{Past}}, {{Present}} and {{Future}},'' \emph{arXiv 1610.02984}, 2016.

\bibitem{chen2017multi}
W.~Chen, X.~Chen, J.~Zhang, and K.~Huang, ``A multi-task deep network for
  person re-identification,'' in \emph{Thirty-{{First AAAI Conference}} on
  {{Artificial Intelligence}}}, 2017.

\bibitem{Quad_Loss}
------, ``Beyond {{Triplet Loss}}: {{A Deep Quadruplet Network}} for {{Person
  Re}}-{{Identification}},'' in \emph{The {{IEEE Conference}} on {{Computer
  Vision}} and {{Pattern Recognition}} ({{CVPR}})}, 2017.

\bibitem{PDC}
C.~Su, J.~Li, S.~Zhang, J.~Xing, W.~Gao, and Q.~Tian, ``Pose-driven {{Deep
  Convolutional Model}} for {{Person Re}}-identification,'' in \emph{{{ICCV}}},
  2017.

\bibitem{SpindleNet}
H.~Zhao, M.~Tian, S.~Sun, J.~Shao, J.~Yan, S.~Yi, X.~Wang, and X.~Tang,
  ``Spindle {{Net}}: {{Person Re}}-{{Identification With Human Body Region
  Guided Feature Decomposition}} and {{Fusion}},'' in \emph{The {{IEEE
  Conference}} on {{Computer Vision}} and {{Pattern Recognition}} ({{CVPR}})},
  2017.

\bibitem{PoseBox}
L.~Zheng, Y.~Huang, H.~Lu, and Y.~Yang, ``Pose {{Invariant Embedding}} for
  {{Deep Person Re}}-identification,'' \emph{IEEE Transactions on Image
  Processing}, 2019.

\bibitem{PN-GAN}
X.~Qian, Y.~Fu, T.~Xiang, W.~Wang, J.~Qiu, Y.~Wu, Y.-G. Jiang, and X.~Xue,
  ``Pose-{{Normalized Image Generation}} for {{Person Re}}-identification,'' in
  \emph{The {{European Conference}} on {{Computer Vision}} ({{ECCV}})}, 2018.

\bibitem{FD-GAN}
Y.~Ge, Z.~Li, H.~Zhao, G.~Yin, S.~Yi, X.~Wang, and H.~Li, ``{{FD}}-{{GAN}}:
  {{Pose}}-guided {{Feature Distilling GAN}} for {{Robust Person
  Re}}-identification,'' in \emph{{{NIPS}}}, 2018.

\bibitem{Siamese-LSTM}
R.~R. Varior, B.~Shuai, J.~Lu, D.~Xu, and G.~Wang, ``A siamese long short-term
  memory architecture for human re-identification,'' in \emph{European
  {{Conference}} on {{Computer Vision}} ({{ECCV}})}, 2016.

\bibitem{PCB+RPP}
Y.~Sun, L.~Zheng, Y.~Yang, Q.~Tian, and S.~Wang, ``Beyond {{Part Models}}:
  {{Person Retrieval}} with {{Refined Part Pooling}},'' in \emph{The {{European
  Conference}} on {{Computer Vision}} ({{ECCV}})}, 2018.

\bibitem{latent_parts}
D.~Li, X.~Chen, Z.~Zhang, and K.~Huang, ``Learning {{Deep Context}}-{{Aware
  Features Over Body}} and {{Latent Parts}} for {{Person
  Re}}-{{Identification}},'' in \emph{The {{IEEE Conference}} on {{Computer
  Vision}} and {{Pattern Recognition}} ({{CVPR}})}, 2017.

\bibitem{STN}
M.~Jaderberg, K.~Simonyan, A.~Zisserman, and {Kavukcuoglu, Koray}, ``Spatial
  transformer networks,'' in \emph{Advances in {{Neural Information Processing
  Systems}}}, 2015.

\bibitem{cluster_parts}
H.~Yao, S.~Zhang, Y.~Zhang, J.~Li, and Q.~Tian, ``Deep {{Representation
  Learning}} with {{Part Loss}} for {{Person Re}}-{{Identification}},''
  \emph{IEEE Transactions on Image Processing}, 2019.

\bibitem{matsukawaPersonReidentificationUsing2016}
T.~Matsukawa and E.~Suzuki, ``Person re-identification using {{CNN}} features
  learned from combination of attributes,'' in \emph{Proceedings of
  {{International Conference}} on {{Pattern Recognition}} ({{ICPR}})}, 2016.

\bibitem{xiaoLearningDeepFeature2016}
T.~Xiao, H.~Li, W.~Ouyang, and X.~Wang, ``Learning {{Deep Feature
  Representations With Domain Guided Dropout}} for {{Person
  Re}}-{{Identification}},'' in \emph{The {{IEEE Conference}} on {{Computer
  Vision}} and {{Pattern Recognition}} ({{CVPR}})}, 2016.

\bibitem{Mask-Guided}
C.~Song, Y.~Huang, W.~Ouyang, and L.~Wang, ``Mask-{{Guided Contrastive
  Attention Model}} for {{Person Re}}-{{Identification}},'' in \emph{The {{IEEE
  Conference}} on {{Computer Vision}} and {{Pattern Recognition}} ({{CVPR}})},
  2018.

\bibitem{Mutual}
Y.~Zhang, T.~Xiang, T.~M. Hospedales, and H.~Lu, ``Deep {{Mutual Learning}},''
  in \emph{The {{IEEE Conference}} on {{Computer Vision}} and {{Pattern
  Recognition}} ({{CVPR}})}, 2018.

\bibitem{SGGNN}
Y.~Shen, H.~Li, S.~Yi, D.~Chen, and X.~Wang, ``Person {{Re}}-identification
  with {{Deep Similarity}}-{{Guided Graph Neural Network}},'' in \emph{The
  {{European Conference}} on {{Computer Vision}} ({{ECCV}})}, 2018.

\bibitem{ResNet}
K.~He, X.~Zhang, S.~Ren, and J.~Sun, ``Deep residual learning for image
  recognition,'' in \emph{The {{IEEE Conference}} on {{Computer Vision}} and
  {{Pattern Recognition}} ({{CVPR}})}, 2016.

\bibitem{Adam}
D.~Kingma and J.~Ba, ``Adam: {{A}} method for stochastic optimization,'' in
  \emph{International {{Conference}} on {{Learning Representations}}
  ({{ICLR}})}, 2015.

\bibitem{t-SNE}
L.~van~der Maaten and G.~Hinton, ``Visualizing data using t-{{SNE}},''
  \emph{Journal of machine learning research}, 2008.

\end{thebibliography}
}

\end{document}